\documentclass{article}

\usepackage{microtype}
\usepackage{graphicx}
\usepackage{subcaption}
\usepackage{booktabs} % for professional tables

\usepackage[accepted]{icml2019}

\icmltitlerunning{World Discovery Models}

\usepackage{amsmath}
\usepackage{amsfonts}
\usepackage{amssymb}
\usepackage{amsthm}
\usepackage[noabbrev,capitalize]{cleveref}
\usepackage{autonum}
\usepackage{siunitx}

\usepackage[english]{babel}
\usepackage{algorithm,algorithmic}
\usepackage{color}
\usepackage[toc,page]{appendix} 

\newcommand{\beq}{\begin{equation}}
\newcommand{\eeq}{\end{equation}}

\newcommand{\beqa}{\begin{eqnarray}}
\newcommand{\eeqa}{\end{eqnarray}}

\newcommand{\beqan}{\begin{eqnarray*}}
\newcommand{\eeqan}{\end{eqnarray*}}

\newcommand{\wh}{\widehat}

\newcommand{\fiverooms}{{\small \texttt{5\,rooms}}}
\newcommand{\maze}{{\small \texttt{maze}}}
\newcommand{\fixedObj}{{\small \texttt{fixed}}}
\newcommand{\bouncingObj}{{\small \texttt{bouncing}}}
\newcommand{\brownianObj}{{\small \texttt{Brownian}}}
\newcommand{\tvObj}{{\small \texttt{white noise}}}
\newcommand{\movableObj}{{\small \texttt{movable}}}

\newlength{\minipagewidth}
\newcommand{\eqdef}{\stackrel{\rm def}{=}}

\newtheorem{predefinition}{Definition}

\newtheorem{theorem}{Theorem}

\newtheorem{preproposition}{Proposition}

\renewcommand{\hat}{\widehat}
\renewcommand{\epsilon}{\varepsilon}

\newcommand{\A}{\mathcal{A}}

\renewcommand{\algorithmicrequire}{\textbf{Input:}}
\renewcommand{\algorithmicensure}{\textbf{Output:}}
\renewcommand{\algorithmiccomment}[1]{\bgroup\hfill\small \texttt{$\triangleright$~#1}\egroup}

\begin{document}

\twocolumn[
\icmltitle{World Discovery Models} % Remi do you like it?

\icmlsetsymbol{equal}{*}

\begin{icmlauthorlist}
\icmlauthor{Mohammad Gheshlaghi Azar}{equal,DM}
\icmlauthor{Bilal Piot}{equal,DM}
\icmlauthor{Bernardo Avila Pires}{equal,DM}
\icmlauthor{Jean-Bastien Grill}{DMP}
\icmlauthor{Florent Altch\'e}{DMP}
\icmlauthor{ R\'emi Munos}{DMP}
\end{icmlauthorlist}

\icmlaffiliation{DM}{DeepMind, London, UK}
\icmlaffiliation{DMP}{DeepMind Paris, France}

\icmlcorrespondingauthor{Mohammad Gheshlaghi Azar}{mazar@google.com}


\icmlkeywords{exploration, information gain, partially observable, reinforcement learning, RL, intrinsic motivation, discovery, curiosity, unsupervised learning}

\vskip 0.3in
]

 \printAffiliationsAndNotice{\icmlEqualContribution} % otherwise use the standard text.

\begin{abstract}
As humans we are driven by a strong desire for seeking novelty in our world. Also upon observing a novel pattern we are capable of refining our understanding of the world based on the new information---humans can \emph{discover} their world. The outstanding ability of the human mind for discovery has led to many breakthroughs in science, art and technology. Here we investigate the possibility of building an agent capable of discovering its world using the modern AI technology. In particular we introduce NDIGO, Neural Differential Information Gain Optimisation, a self-supervised discovery model that aims at seeking new information to construct a global view of its world from partial and noisy observations. Our experiments on some controlled 2-D navigation tasks show that NDIGO outperforms state-of-the-art information-seeking methods in terms of the quality of the learned representation. The improvement in performance is particularly significant in the presence of white or structured noise where other information-seeking methods follow the noise instead of discovering their world.
\end{abstract}
\section{Introduction}
\label{sec:introduction}
Modern AI has been remarkably successful in solving complex decision-making problems such as GO~\citep{silver2016mastering, silver2017mastering}, simulated control tasks~\citep{schulman2015trust}, robotics~\citep{levine2016end}, poker~\citep{moravvcik2017deepstack} and Atari games~\citep{mnih2015human,hessel2018rainbow}. Despite these successes the agents developed by those methods are specialists: they perform extremely well at the tasks they were trained on but are not very successful at generalising their task-dependent skills in the form of a general domain understanding. Also, the success of the existing AI agents often depends strongly on the availability of external feedback from their world in the form of reward signals or labelled data, for which some level of supervision is required. This is in contrast to the human mind, which is a general and self-supervised learning system that \emph{discovers} the world around it even when no external reinforcement is available. Discovery is the ability to obtain knowledge of a phenomenon for the first time~\citep{merriam2004merriam}. As discovery  entails the process of learning of and about new things, it is  an integral part of what makes humans capable of understanding their world in a task-independent and self-supervised fashion.

The underlying process of discovery in humans is complex and multifaceted~\citep{hohwy2013predictive}. However one can identify two main mechanisms for discovery~\citep{clark2017nice}. The first mechanism is  {\bf active information seeking}. One of the primary behaviours of humans is their attraction to  novelty (new information) in their world~\citep{litman2005curiosity,kidd2015psychology}. The human mind is very good at distinguishing between the \emph{novel} and the \emph{known}, and this ability is partially due to the extensive internal reward mechanisms of \emph{surprise}, \emph{curiosity}  and \emph{excitement}~\citep{schmidhuber2009simple}. The second mechanism is {\bf building a statistical world model}. Within cognitive neuroscience, the theory of statistical predictive mind states that the brain, like scientists, constructs and maintains a set of hypotheses over its representation of the world~\citep{friston2014computational}. Upon perceiving a novelty, our brain has the ability to validate the existing hypothesis, reinforce the ones which are compatible with the new observation and discard the incompatible ones. This self-supervised process of hypothesis building is essentially how humans consolidate their ever-growing knowledge in the form of an accurate and global model.                              
Inspired by these inputs from cognitive neuroscience, information-seeking algorithms have received significant attention to improve the exploration capability of artificial learning agents~\citep{schmidhuber1991possibility, houthooft2016vime,achiam2017surprise,pathak2017curiosity,burda2018large,shyam2018model}.  However, the scope of the existing information-seeking algorithms is often limited to the case of fully observable and deterministic environments. One of the problems with the existing novelty-seeking algorithms is that agents trained by these methods tend to become attracted to random patterns in their world and stop exploring upon encountering them, despite the fact that these random patterns contain no actual \emph{information} on the world~\citep{burda2018large}. Moreover, the performance of existing agents are often evaluated based on their ability to solve a reinforcement learning (RL) task with extrinsic reward, and not on the quality of the learned world representation, which is the actual goal of discovery. Thus, it is not clear whether the existing algorithms are capable of using the novel information to discover their world. Therefore, the problem of discovery in the general case of partially observable and stochastic environments remains open.

The main contribution of this paper is to develop a practical and end-to-end algorithm for discovery in stochastic and partially observable worlds using modern AI technology. We achieve this goal by designing a simple yet effective algorithm called NDIGO, {\bf N}eural {\bf D}ifferential {\bf I}nformation {\bf G}ain {\bf O}ptimisation, for information seeking designed specifically for stochastic partially observable domains.  NDIGO identifies novelty by measuring the increment of information provided by a new observation in predicting the future observations, compared to a baseline prediction for which this observation is withheld. We show that this measure can be estimated using the difference of prediction losses of two estimators, one of which can access the complete set of observations while the other does not receive the latest observation. We then use this measure of novelty as the intrinsic reward to train the policy using a state of the art reinforcement learning algorithm~\citep{kapturowski2018recurrent}. One of the key features of NDIGO is its robustness to noise, as the process of subtracting prediction losses cancels out errors that the algorithm cannot improve on. Moreover, NDIGO is well-suited for discovery in partially observable domains as the measure of novelty in NDIGO drives the agent to the unobserved areas of the world where new information can be gained from the observations. Our experiments show that NDIGO produces a robust performance in the presence of noise in partial observable environments: NDIGO not only finds true novelty without being distracted by the noise, but it also incorporates this information into its world representation without forgetting previous observation. 

\section{Related Work}
\label{sec:related works}
It has been argued for decades in developmental psychology~\citep{white1959motivation, deci1985intrinsic, csikszentmihalyi1992optimal}, neuroscience~\citep{dayan2002reward, kakade2002dopamine, horvitz2000mesolimbocortical} and machine learning~\citep{oudeyer2008can, gottlieb2013information, schmidhuber1991curious} that an agent maximising a simple intrinsic reward based on patterns that are both novel and learnable could exhibit essential aspects of intelligence such as autonomous development~\cite{oudeyer2016evolution}.

More specifically, in his survey on the theory of creativity and intrinsic motivation, \citet{schmidhuber2010formal} explains how to build the agent that could discover and understand in a self-supervised way its environment. He establishes that $4$ crucial components are necessary: {\bf i)} a world model~\citep{ha2018world} that encodes what is currently known. It can be a working memory component such as a Long Short Term Memory network~\citep[LSTM,][]{hochreiter1997long} or a Gated Recurrent Unit network~\citep[GRU,][]{cho2014learning}. {\bf ii)} a learning algorithm that improves the world model. For instance, \citet{guo2018} have shown that a GRU trained with a Contrastive Prediction Coding \citep[CPC,][]{oord2018representation} loss on future frames could learn a representation of the agent's current and past position and orientation, as well as position of objects in the environment. {\bf iii)} An intrinsic reward generator based on the world model that produces rewards for patterns that are both novel and learnable. Different types of intrinsic rewards can be used, such as the world model's prediction error~\citep{stadie2015incentivizing, pathak2017curiosity}, improvement of the model's prediction error, also known as prediction gain~\citep{achiam2017surprise, schmidhuber1991curious, lopes2012exploration}, and finally information gain~\citep{shyam2018model, itti2009bayesian, little2013learning, frank2014curiosity, houthooft2016vime}. {\bf iv} the last component is an RL algorithm that finds an optimal policy with respect to the intrinsic rewards.

Recently, several implementations of intrinsically motivated agents have been attempted using modern AI technology. Most of them used the concept of prediction error as an intrinsic reward~\citep{stadie2015incentivizing,pathak2017curiosity,burda2018large, haber2018learning}. However, it has been argued that agents optimising the prediction error are susceptible to being attracted to white noise~\citep{oudeyer2007intrinsic} and therefore should be avoided. To solve the white-noise problem, different types of random or learned projections~\citep{burda2018large} of the original image into a smaller feature space less susceptible to white-noise are considered. Other implementations rely on approximations of the concept of information gain~\citep{houthooft2016vime, achiam2017surprise} via a variational lower bound argument. Indeed, as they are trying to train a probabilistic model over the set of possible dynamics, the computation of the posterior of that distribution is intractable~\citep{houthooft2016vime}. Finally, models based on prediction gain are fundamentally harder to train compared to prediction error~\cite{achiam2017surprise,lopes2012exploration, pathak2017curiosity, ostrovski2017count}.  Also it is not entirely clear  how effective they are in seeking novelty  in comparison with  methods that rely on information gain~\cite{schmidhuber2010formal}. 

\section{Setting}
\label{sec:general setting}
We consider a partially observable environment where an agent is shown an observation $o_t$ at time $t$, then selects an action $a_t$ which generates a new observation $o_{t+1}$ at the next time step. We assume observations $o_t$ are generated by an underlying process $x_t$ following Markov dynamics, i.e.~$x_{t+1} \sim P(\cdot |x_t,a_t)$, where $P$ is the dynamics of the underlying process. Although we do not explicitly use the corresponding terminology, this process can be formalised in terms of Partially Observable Markov Decision Processes~\citep[POMDPs;][]{lovejoy1991survey,cassandra1998exact}.

The future observation $o_{t+1}$ in a POMDP can also be seen as the output of a stochastic mapping with input the current history. Indeed, at any given time $t$, let the current history $h_t$ be all past actions and observations $h_t \eqdef (o_0, a_0, o_1, a_1, \cdots, a_{t-1}, o_t)$. Then we define $\mathbb{P}(\cdot|h_{t},a_t)$ the probability distribution of $o_t$ knowing the history and the action $a_t$. One can generalise this notion for k-step prediction: for any integers $t\geq0$ and $k\geq1$, let us denote by $t:t+k$ the integer interval $\{t,\dots, t+k-1\}$, and let $\A_{t:t+k}\eqdef (a_t,\dots,a_{t+k-1})$  and $\mathcal O_{t:t+k}\eqdef (o_t,\dots,o_{t+k-1})$ be the sequence of actions and observations from time $t$ up to time $t+k-1$, respectively. Then $o_{t+k}$ can be seen as a sample drawn from the probability distribution $\mathbb{P}(\cdot|h_{t}, \A_{t:t+k})$, which is the $k$-step open-loop prediction model of the observation $o_{t+k}$. We also use the short-hand notation $\mathbb{P}_{t+k|t}=\mathbb{P}(\cdot|h_{t}, \A_{t:t+k})$ as the probability distribution  of $o_{t+k}$ given the history $h_t$ and the sequence of actions $\A_{t:t+k}$. 

\section{Learning the World Model}
The world model should capture what the agent currently knows about the world so that he could make predictions  based on what it knows. We thus build a model of the world by predicting future observations given the past~\citep[see e. g.,][]{schmidhuber1991curious, guo2018}. More precisely, we build an internal representation $b_t$ by making predictions of futures frames $o_{t+k}$ conditioned on a sequence of actions $\A_{t:t+k}$ and given the past $h_t$. This is similar to the approach of Predictive State Representations \citep[]{littman2002predictive}, from which we know that if the learnt representation $b_t$ is able to predict the probability of any future observation conditioned on any sequence of actions and history, then this representation $b_t$ contains all information about the belief state (i.e., distribution over the ground truth state $x_t$). 

\subsection{Architecture}
We propose to learn the world model by using a recurrent neural network (RNN) $f_\theta$ fed with the concatenation of observation features $z_t$ and the action $a_t$ (encoded as a one-hot vector). The observation features $z_t$ are obtained by applying a convolutional neural network (CNN) $f_{\phi}$ to the observation $o_t$. The RNN is a Gated Recurrent Unit (GRU) and the internal representation is the hidden state of the GRU, that is, $b_t = f_\theta(z_t, a_{t-1}, b_{t-1})$, as shown in \Cref{fig:worldmodel}. We initialise this GRU by setting its hidden state to the null vector $0$, and using $b_0 = f_\theta(z_0, a, 0)$ where $a$ is a fixed, arbitrary action and $z_0$ are the features corresponding to the original observation $o_0$. We train this representation $b_t$ with some future-frame prediction tasks conditioned on sequences of actions and the representation $b_t$. These frame prediction tasks consist in estimating the probability distribution, for various $K\geq k\geq1$ (with $K\in\mathbb{N}^*$ to be specified later), of future observation $o_{t+k}$ conditioned on the internal representation $b_t$ and the sequence of actions $\A_{t:t+k}$. We denote these estimates by $\hat{p}_{t+k|t}(.|b_t, \A_{t:t+k})$ or simply by $\hat{p}_{t+k|t}$ for conciseness and when no confusion is possible. As the notation suggests, we will use $\hat{p}_{t+k|t}$ as an estimate of $\mathbb{P}_{t+k|t}$.
The neural architecture consists in $K$ different neural nets $\{f_{\psi_k}\}_{k=1}^K$. Each neural net $f_{\psi_k}$ receives as input the concatenation of the internal representation $b_t$ and the sequence of actions $\A_{t:t+k}$, and outputs the distributions over observations: $\hat{p}_{t+k|t} = f_{\psi_k}(b_t, \A_{t:t+k})$ (). For a fixed $t\geq0$ and a fixed $K\geq k\geq1$, the loss function $L(o_{t+k}, \hat{p}_{t+k|t})$ at time step $t+k-1$ associated with the network $f_{\psi_k}$ is a cross entropy loss: $L(o_{t+k}, \hat{p}_{t+k|t})= - \ln (\hat{p}_{t+k|t}(o_{t+k})).$ We finally define for any given sequence of actions and observations the \emph{representation loss} function $L_{\text{repr}}$ as the sum of these cross entropy losses: $L_{\text{repr}} = \sum_{t\geq 0, K\geq k\geq1} L(o_{t+k}, \hat{p}_{t+k|t})$. 
\begin{centering}
\begin{figure}[ht!]\centering
\includegraphics[width=.45\textwidth]{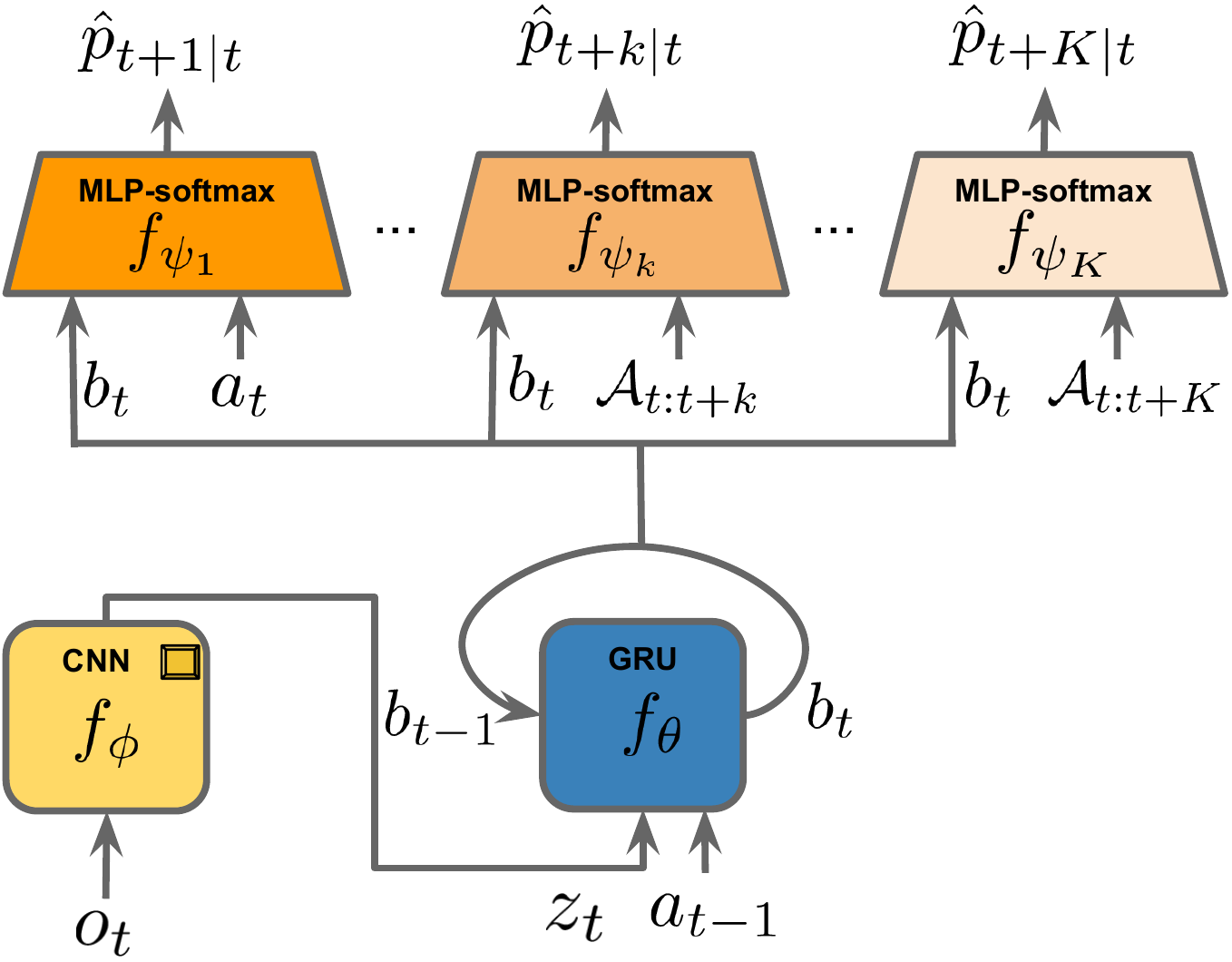}
   \caption{World Model: a CNN and a GRU encode the history $h_{t}$ into an internal representation $b_t$. Then, $K$ frame predictions tasks are trained in order to shape the representation $b_t$.}
    \label{fig:worldmodel}
\end{figure}+
\end{centering} 
\subsection{Evaluation of the learnt representation} 
\label{subsec:evaluation}
In the POMDP setting, the real state $x_t$ represents all there is to know about the world at time $t$.  By constructing a belief state, which is a distribution $P_{b}(\cdot|h_t)$ over the possible states conditioned on the history $h_t$, the agent can assess its uncertainty about the real state $x_t$ given the history $h_t$. Therefore, in order to assess the quality of the learnt representation $b_t$, we use the glass-box approach described in \Cref{fig:globalNDIGOarchitecture} to build a belief state of the world. It consists simply in training a neural network $f_{\tau}$ fed by the internal representation $b_t$ to predict a distribution $\hat{p}_b(\cdot|b_t)$ over the possible real state $x_t$. This kind of approach is only possible in artificial or controlled environments where the real state is available to the experimenter but yet not given to the agent. We also make sure that no gradient from $f_{\tau}$ is being back-propagated to the internal representation $b_t$ such that the evaluation does not influence the learning of the representation and the behaviour of the agent. For a fixed $t\geq0$,  the loss used to trained $f_{\tau}$ is a cross entropy loss \citep[For a more detailed description of the approach see][]{guo2018}: $L_{\text{discovery}}(x_{t}, \hat{p}_b(\cdot|b_t))\eqdef - \ln (\hat{p}_b(x_{t}|b_t))$.
We call this loss \emph{discovery loss}, and use it as a measure of how much information about the whole world the agent is able to encode in its internal representation $b_t$, i.e., how much of the world  has been discovered by the agent.

\section{NDIGO Agent}
\label{sec:agent architecture}

Our NDIGO  agent is a discovery agent that learns to seek new information in its environment and then incorporate this information into a world representation. Inspired by the intrinsic motivation literature~\citep{schmidhuber2010formal}, the NDIGO  agent achieves this information-seeking behaviour as a result of optimising an intrinsic reward. Therefore, the agent's exploratory skills depend critically on designing an appropriate reward signal that encourages discovering the world. Ideally, we want this reward signal to be high when the agent gets an observation containing new information about the real state $x_t$. As we cannot access $x_t$ at training time, we rely on  the accuracy of our future observations predictions to estimate the information we have about $x_t$.

Intuitively, for a fixed horizon $H\in\mathbb{N}^*$, the \emph{prediction error loss} $L(o_{t+H}, \hat{p}_{t+H|t}) = -\ln (\hat{p}_{t+H|t}(o_{t+H}))$ is a good measure on how much information $b_t$ is lacking about the future observation $o_{t+H}$. The higher the loss, the more uncertain the agent is about the future observation $o_{t+H}$ so the less information it has about this observation. Therefore, one could define an intrinsic reward directly as the prediction error loss, thus encouraging the agent to move towards states for which it is the less capable of predicting future observations. The hope is that the less information we have in a certain belief state, the easier it is to gain new information. Although this approach may have good results in deterministic environments, it is however not suitable in certain stochastic environments. For instance, consider the extreme case in which the agent is offered to observe white noise such as a TV displaying static. An agent motivated with prediction error loss would continually receive a high intrinsic reward simply by staying in front of this TV, as it cannot improve its predictions of future observations, and would effectively remain fascinated by this noise.

\subsection{The NDIGO intrinsic reward}
\label{sec:intrinsic}
The reason why the naive prediction error reward fails in such a simple example is that the agent identifies that a lot of information is lacking, but does not acknowledge that no progress is made towards acquiring this lacking information. To overcome this issue, we introduce the NDIGO reward, for a fixed $K\geq H\geq 1$, as follows: 
\begin{equation}
\label{eq:NDIGO}
    r_{t+H-1}^{\mathrm{NDIGO}} \eqdef L(o_{t+H}, \wh{p}_{t+H|t-1}) - L(o_{t+H}, \wh{p}_{t+H|t}),
\end{equation}
where $o_{t+H}$ represents the future observation considered and $H$ is the horizon of NDIGO. The two terms in the right-hand side of \cref{eq:NDIGO} measure how much information the agent lacks about the future observation $o_{t+H}$ knowing all past observations prior to $o_t$ with $o_t$ either excluded (left term) or included (right term). Intuitively, we take the difference between the information we have at time $t$ with the information we have at time $t-1$. This way we get an estimate of how much information the agent gained about $o_{t+H}$ by observing $o_t$. Note that the reward $r_{t+H-1}^\mathrm{NDIGO}$ is attributed at time $t+H-1$ in order to make it dependent on $h_{t+H-1}$ and $a_{t+H-1}$ only (and not on the policy), once the prediction model $\hat p$ has been learnt. If the reward had been assigned at time $t$ instead (time of prediction) it would have depended on the policy used to generate the action sequence $\A_{t:t+H-1}$, which would have violated the Markovian assumption required to train the RL algorithm. Coming back to our broken TV example, the white noise in $o_{t}$ does not help in predicting the future observation $o_{t+H}$. The NDIGO reward is then the difference of two large terms of similar amplitude, leading to a small reward: while acknowledging that a lot of information is missing (large prediction error loss) NDIGO also realises that no more of it can be extracted (small difference of prediction error loss). Our experiments show that using NDIGO allows the agent to avoid being stuck in the presence of noise, as presented in \Cref{sec:experiments}, thus confirming these theoretical considerations.

\subsection{Algorithm}
\label{sec:ndigo_algo}
Given the intrinsic reward $r_{t+H-1}^\mathrm{NDIGO}$,  we use the state-of-the-art RL algorithm R2D2~\citep{kapturowski2018recurrent} to optimise the policy. The NDIGO agent interacts with its world using the NDIGO policy to obtain new observation $o_{t+k}$, which is used to train the world model by minimising the future prediction loss $L_{t+k|t}=L(o_{t+k}, \hat{p}_{t+k|t})$. The losses $L_{t+k|t}$ are then used to obtain the intrinsic reward at the next time step, and the process is then repeated. An in-depth description of the complete NDIGO algorithm can be found in \Cref{appendix:ndigo-alg}.

\subsection{Relation to information gain}
\label{sec:infogain_relation}
Information gain has been widely used as the  novelty signal in the literature~\citep{houthooft2016vime,little2013learning}. A very broad definition of the information gain~\citep{schmidhuber2010formal} is the distance (or divergence) between distributions on any  random event of interest $\omega$ before and after a new sequence of observations. Choosing the random event to be the future observations or actions and the divergence to be the Kullback-Leiber divergence then the $k$-step predictive information gain $IG(o_{t+k}, \mathcal O_{t:t+k}|h_t, \A_{t:t+k})$ of the future event $o_{t+k}$ with respect to the sequence of observations $\mathcal O_{t:t+k}$ is defined as: $IG(o_{t+k}, \mathcal O_{t:t+k}|h_t, \A_{t:t+k}) \eqdef \mathrm{KL}\left(\mathbb{P}_{t+k|t+k-1} || \mathbb{P}_{t+k|t-1} \right)$, and measures how much information can be gained about the future observation $o_{t+k}$ from the sequence of past observations $\mathcal O_{t:t+k}$ given the whole history $h_{t}$ up to time step $t$ and the sequence of actions $\A_{t:t+k}$ from $t$ up to $t+H-1$. In the case of  $k=1$ we recover the 1-step information gain on the next observation $o_{t+1}$ due to $o_t$. We also use the following short-hand notation for the information gain $IG_{t+k|t} = IG(o_{t+k}, \mathcal O_{t:t+k}|h_t, \A_{t:t+k})$ for every $k\geq 1$ and $t\geq0$. Also by convention we define $IG_{t|t}=0$. 

We now show that the NDIGO intrinsic reward $r_{t+H-1}^\mathrm{NDIGO}$  can be  expressed as the difference of information gain due to $\mathcal O_{t:t+H}$ and $\mathcal O_{t+1:t+H}$. For a given horizon $H\geq1$ and $t\geq0$, the intrinsic reward for time step $t+H-1$ is:
\vspace{-4pt}
\begin{align}
r^{\mathrm{NDIGO }}_{t+H-1} & \eqdef L(o_{t+H}, \hat{p}_{t+H|t-1}) - L(o_{t+H}, \hat{p}_{t+H|t})
\\
& = \ln\left(\frac{\hat{p}_{t+H|t}(o_{t+H})}{\hat{p}_{t+H|t-1}(o_{t+H})}\right).
\end{align}
\vspace{-2pt}
Given that $\hat{p}_{t+H|t}$ and $\hat{p}_{t+H|t-1}$ are respectively an estimate of $\mathbb{P}_{t+H|t}$ and  $\mathbb{P}_{t+H|t-1}$, and based on the fact that these estimates become more accurate as the number of samples increases, we have:
\vspace{-2pt}
\begin{align}
\mathbb{E}\left[ r_{t+H-1}^\mathrm{NDIGO} \right] &= \mathbb{E}_{o_{t+H}\sim\mathbb{P}_{t+H|t+H-1}}\left[ \ln\left(\frac{\hat{p}_{t+H|t}(o_{t+H})}{\hat{p}_{t+H|t-1}(o_{t+H})}\right)\right]\notag
\\
& \approxeq  \mathbb{E}_{o_{t+H}\sim\mathbb{P}_{t+H|t+H-1}}\left[ \ln\left(\frac{\mathbb{P}_{t+H|t}(o_{t+H})}{\mathbb{P}_{t+H|t-1}(o_{t+H})}\right)\right]\notag
\\
&=\mathrm{KL}(\mathbb{P}_{t+H|t+H-1}||\mathbb{P}_{t+H|t-1})\notag
\\
&\hspace{2.2cm} -\mathrm{KL}(\mathbb{P}_{t+H|t+H-1}||\mathbb{P}_{t+H|t})\notag
\\
&=
IG_{t+H|t} - IG_{t+H|t-1}\label{eq:delta-gain}.
\end{align}
\vspace{-2pt}
The first term $IG_{t+H|t}$ in \cref{eq:delta-gain} measures how much information can be gained about $o_{t+H}$ from the sequence of past observations $\mathcal O_{t:t+H}$ whereas the second term $IG_{t+H|t+1}$  measures how much information can be gained about $o_{t+H}$  from the sequence of past observations $\mathcal O_{t+1:t+H}$. Therefore, as $\mathcal O_{t+1:t+H}=\mathcal O_{t:t+H}\backslash\{o_t\}$ , the expected value of the NDIGO reward at step $t+H-1$ is equal to the amount of additional information that can be gained by the observation $o_t$ when trying to predict $o_{t+H}$.

\section{Experiments}
\label{sec:experiments}
%\paragraph{Section introduction.}
We evaluate the performance of NDIGO qualitatively and quantitatively on five experiments, where we demonstrate different aspects of discovery with NDIGO.
In all experiments there are some hidden objects which the agent seeks to discover. However the underlying dynamics of the objects are different. In the simplest case, the location of objects only changes at the beginning of every episode, whereas in the most complex the objects are changing their locations throughout the episode according to some random walk strategy. We investigate {\bf (i)} whether the agent can efficiently search for novelty, i.e., finding the location of objects; {\bf (ii)} whether the agent can encode the information of object location in its representation of the world such that the discovery loss of predicting the objects is as small as possible.

\subsection{Baselines}
We compare our algorithm NDIGO-$H$, with $H$ being the horizon and taking values in $\{1,2,4\}$, to different information seeking and exploration baselines considered to be state of the art in the intrinsic motivation literature. {\bf  Prediction Error (PE)} \citep{haber2018learning,achiam2017surprise}: The PE model uses  the same architecture and the same losses than NDIGO. The only difference is that the intrinsic reward is the predictor error: $r^{\mathrm{PE}}_{t} = L(\hat{p}_{t+1|t}, o_{t+1})$. {\bf Prediction Gain (PG)} \citep{achiam2017surprise, ostrovski2017count}: Our version of PG uses the same architecture and the same losses than NDIGO. In addition, at every $n=2$ learner steps we save a copy of the prediction network into a fixed target network. The intrinsic reward is the difference in prediction error, between the up-to-date network and the target network predictions: $r^{\mathrm{PG}}_{t} = L(\hat{p}^{\mathrm{target}}_{t+1|t}, o_{t+1} ) - L(\hat{p}_{t+1|t}, o_{t+1} )$, where $\hat{p}^{\mathrm{target}\vphantom |}_{t+1|t}$ is the distribution computed with the weights of the fixed target network. {\bf Intrinsic Curiosity Module (ICM)}~\citep{pathak2017curiosity, burda2018large}: The method consists in training the internal representation $b_t$ to be less sensitive to noise using a self-supervised inverse dynamics model. Then a forward model is used to predict the future internal representation $\hat{b}_{t+1}$ from the actual representation $b_t$ and the action $a_t$ (more details on this model are in \cref{appendix:icm}). The intrinsic reward $r^{\mathrm{FPE}}_{t} = \left\|\hat{b}_{t+1}-b_{t+1}\right\|^2_2$.

\subsection{Test environments} 
\begin{figure}[ht!]
\begin{centering}
    \includegraphics[width=.7\columnwidth]{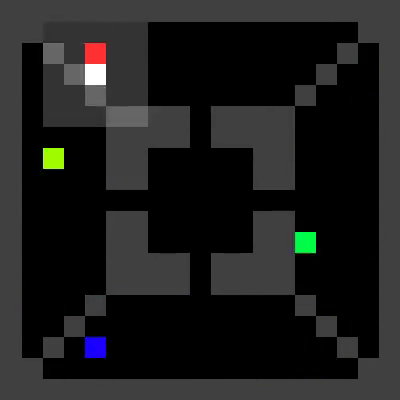}
    \vspace*{-1ex}
    \caption{ The \fiverooms{} environment: in this instance, we can see in white the agent, $4$ fixed objects in each of the $4$ peripheral rooms and in grey the impenetrable walls. The shaded area around the agent represents its $5 \times 5$ region-cell local view.}
    \label{fig:fiverooms}
\end{centering}     
\end{figure}

\paragraph{The \fiverooms{} environment.}
The \fiverooms{} environment (see \cref{fig:fiverooms}) is a local-view 2D environment composed of $5$ rooms implemented using the \texttt{\small pycolab} library\footnote{\texttt{https://github.com/deepmind/pycolab}}. In \texttt{\small pycolab}, the environment is composed of cells that contain features such as walls, objects or agents. In the \fiverooms{} environment, there is one central $5 \times 5$ room and four peripheral rooms (composed of $48$ cells) that we will refer to as upper, lower, left and right rooms. Each of the four peripheral rooms may contain different types of ``objects'' that occupy a cell exclusively. At every episode, the agent starts in the middle of the central room and the starting position of each object is randomised. The objects may or may not move, but as a general rule in any episode they never leave the room they started in. Finally, we only place objects in the peripheral rooms, and in each room there is never more than one object.

\begin{figure}[ht!]
\begin{centering}
    \includegraphics[width=.8\columnwidth]{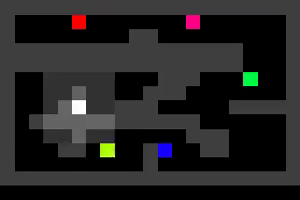}
    \vspace*{-1ex}
    \caption{The \maze{} environment: in this instance, we can see in white the agent, $4$ \fixedObj objects in blue, green, pink and  red.  \tvObj is the closest object to the agent location also in green.}
    \label{fig:maze}
\end{centering}     
\end{figure}

\paragraph{The \maze{} environment.}
The \maze{} environment (see \cref{fig:maze}) is also a \texttt{\small pycolab} local-view 2D environment. It is set-up as a maze composed of six different rooms connected by corridors. The agent starts at a fixed position in the environment in an otherwise empty room $0$; rooms are numbered from $0$ to $5$ based on the order in which they can be reached, i.e. the agent cannot reach room number $3$ without going through rooms $1$ and $2$ in this order. A \tvObj{} object is always present in room $1$, and a there is single \fixedObj{}in rooms $2, 3$ and $4$. Room $5$ contains a special \movableObj{}, which should attract the agent even when the environment is completely learned.


\paragraph{Objects.} 
We consider five different types of objects: \fixedObj{}, \bouncingObj{}, \brownianObj{}, \tvObj{} and \movableObj{}. 
\fixedObj{} objects are fixed during episodes, but change position from episode to episode. 
They provide information gain about their position when it is not already encoded in the agent's representation.
\bouncingObj{} objects bounce in a straight line from wall to wall inside a room.
In addition to providing information gain similar to \fixedObj{} objects, they allow us to test the capacity of the representation to encode predictable object after the object is no longer in the agent's view.
\brownianObj{} objects follow a Brownian motion within a room, by moving uniformly at random in one of the four directions.
\tvObj{} objects change location instantly to any position inside the same room, uniformly at random, at each time step, and are therefore unpredictable.
Finally, \movableObj{} objects do not move by themselves, but the agent can cause them to move to a random location by attempting to move into their cells. Interacting with these objects allows more information gain to be generated.

\paragraph{Agent's observations and actions.} The observation $o_t$ at time $t$ consists in a concatenation of images (called channels) of $25$ pixels representing the different features of the $5 \times 5$ local view of the agent. This can be represented by multidimensional array $(5, 5 , c)$ where $c$ is the number of channels. The first channel represents the walls in the local view: $1$ indicates the presence of a wall and $0$ the absence of a wall. Then, each of the remaining channels represents the position of an object with a one-hot array if the object is present in the local view or with a null array otherwise. The possible actions $a_t$  are stay, up, down, right, left and are encoded with a one-hot vector of size $5$.

\subsection{Performance evaluation} 
The agent's performance is  measured by its capacity to estimate the underlying state of the world from its internal representation (discovery loss, see \cref{subsec:evaluation}). In \texttt{\small{pycolab}}, it is possible to compute a discovery loss for each aspect of the world state (location of each object for instance). So that it is easy to understand which aspects of the world the agent can understand and keep in its internal representation. Once again we stress the fact that no gradient is back-propagated from that evaluation procedure to the internal representation. In addition, we provide other statistics such as average values of first-visit time and visit counts of a given object to describe the behavior of the agent. The first-visit time is the number of episode time steps the agent needs before first observing a given object; the visit count is the total number of time steps where the agent observes the object. Finally, we also provide more qualitative results with videos of the agent discovering the worlds (see
\texttt{https://www.youtube.com/channel/}
\\
\texttt{UC5OPHK7pvsZE-jVclZMvhmQ}).

\subsection{Experimental results}

In this section we evaluate the performance of NDIGO on some controlled navigation task (for the implementation details and the specification of the prediction and  policy networks and the training algorithms  see  \cref{sec:agentDetails}).

\paragraph{Experiment 1.} We evaluate the discovery skills of NDIGO by testing how effectively it can ignore the white noise, from which there is nothing to learn, and discover the location of the \fixedObj{} object. Here, we use a \fiverooms{} setting with a \fixedObj{} object in the upper room, and a \tvObj{} object in the lower room. 
\begin{figure}
    \centering
    \includegraphics[width=\columnwidth]{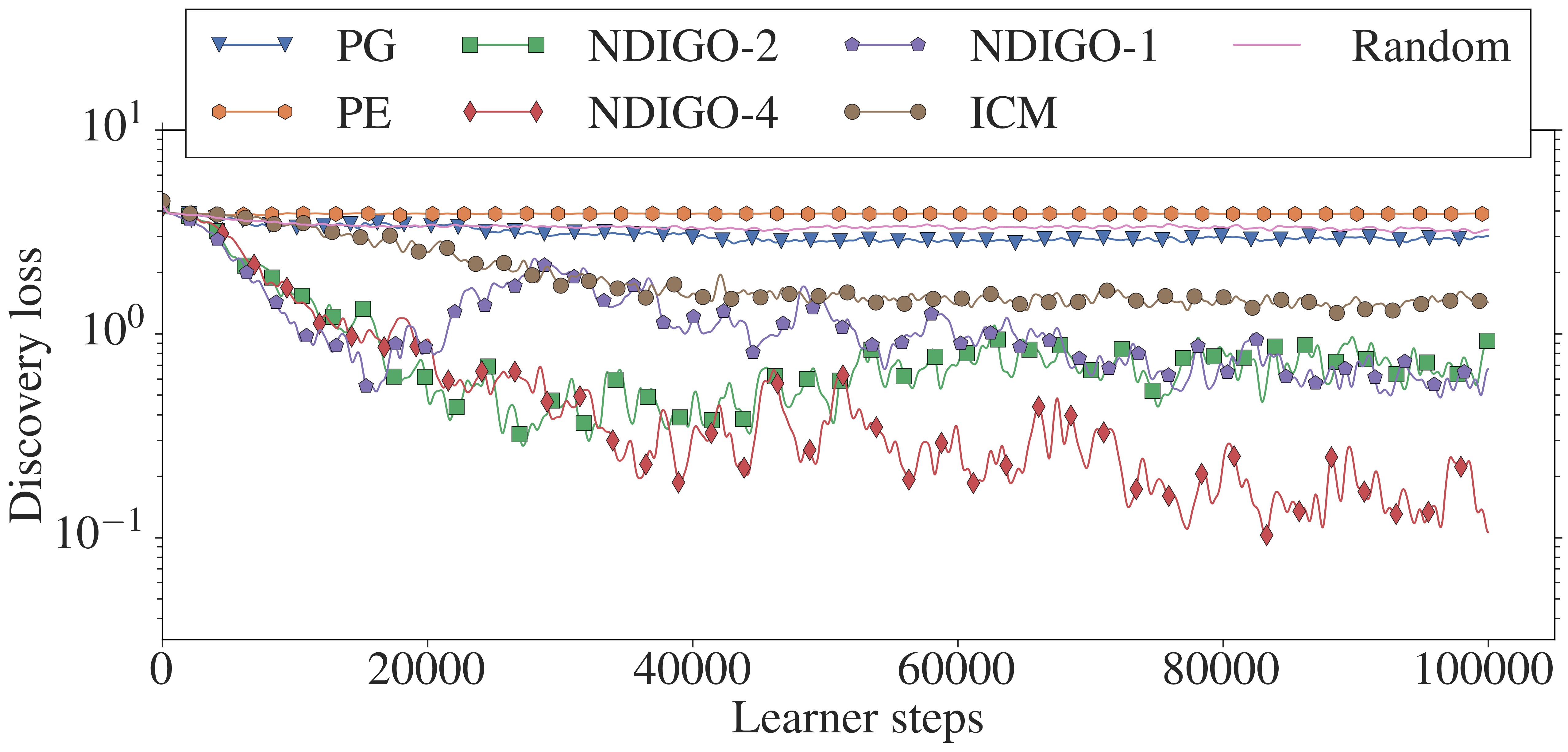}

    \caption{Experiment 1: Average discovery loss of the \fixedObj{} object. The results are averaged over 10 seeds.
    \label{fig:exp1}}

\end{figure}

\begin{table}[ht!]
\scalebox{0.75}{
\begin{tabular}{lcccc}
\toprule 
         & \multicolumn{2}{c}{Visit count} & \multicolumn{2}{c}{First visit time} \\
         & \fixedObj{}   & \texttt{w.~noise}  & \fixedObj{}   & \texttt{w.~noise}  \\
\midrule
Random     &     $14.1 \pm 14.3$     &     $24.6 \pm 12.6$     &     $339.0 \pm 40.5$     &     $225.6 \pm 50.4$ \\
PE     &     $0.1 \pm 0.2$     &     $158.3 \pm 3.7$     &     $392.6 \pm 18.1$     &     $15.5 \pm 4.0$ \\
PG     &     $27.3 \pm 22.0$     &     $22.5 \pm 10.3$     &     $306.4 \pm 49.4$     &     $233.7 \pm 56.6$ \\
ICM     &     $144.8 \pm 37.2$     &     $23.8 \pm 12.4$     &     $132.4 \pm 41.2$     &     $238.3 \pm 55.0$ \\
NDIGO-1     &     $120.9 \pm 43.4$     &     $19.1 \pm 9.3$     &     $78.4 \pm 28.5$     &     $279.4 \pm 42.9$ \\
NDIGO-2     &     $154.0 \pm 45.5$     &     $7.4 \pm 6.7$     &     $112.6 \pm 46.2$     &     $\mathbf{345.8} \pm 36.5$ \\
NDIGO-4     &     $\mathbf{300.4} \pm 22.2$     &     $\mathbf{1.4} \pm 1.2$     &     $\mathbf{40.8} \pm 9.7$     &     $330.7 \pm 47.4$ \\
\bottomrule
\end{tabular}}

\caption{Experiment 1: Average values of the visit counts and first visit time of the trained agent for the \fixedObj{} and \tvObj{} objects in one episode.}
\label{tab:exp1}

\end{table}

We  report in~\cref{fig:exp1} the learning curves for the discovery loss of the \fixedObj{} object. This result shows the quality of the learned representation in terms of encoding  the location of \fixedObj{} object. We observe that the long-horizon variant of NDIGO (NDIGO-4) outperforms the best baseline (ICM) by more than an order of magnitude. Also the asymptotic performance of NDIGO-4 is significantly better than NDIGO-1 and NDIGO-2.     

In \cref{tab:exp1} we also report the average value and standard deviation of visit count and first visit time of the trained agents for the \fixedObj{} object and the \tvObj{} object in an episode\footnote{
Each episode is set to end after $400$ time steps; if an agent does not find the object by the end of the episode, the first visit time is set to $400$.
}. We observe that different variants of NDIGO  are driven towards the \fixedObj{} object and manage to find it faster than the baselines while avoiding the \tvObj{} object. While ICM is also attracted by the \fixedObj{} object, it is not doing it as fast as NDIGO.  PE, as expected, is only attracted by the \tvObj{} object where its reward is the highest. We also observe that the performance of NDIGO improves as we increase the prediction horizon. From now on, in the tables, we report only the ICM results as it is the only competitive baseline. Exhaustive results are reported in~\cref{subsec:additionalexp2-4}.

\paragraph{Experiment 2.} To demonstrate better the information-seeking behaviour of our algorithm, we place randomly a \fixedObj{} object in either the upper, left or right room and a \tvObj{} object in the lower room. Thus, to discover the object, the agent must actively look for it in all but the lower room.

\begin{figure}
    \centering
    \includegraphics[width=\columnwidth]{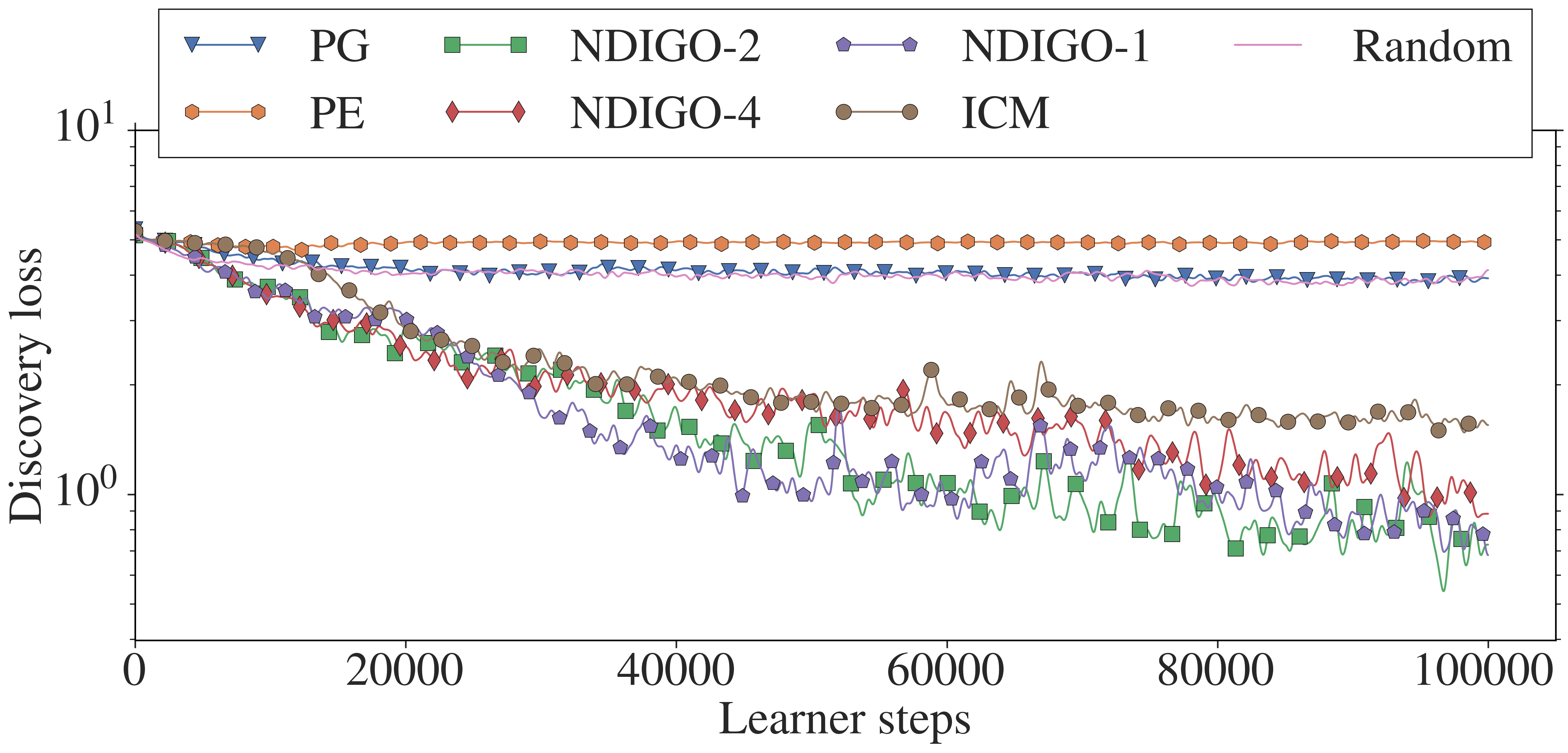}
    \caption{Experiment 2: Average discovery loss of the \fixedObj{} object. The results are averaged over 10 seeds.
    \label{fig:exp2}}
\end{figure}

Similar to Experiment 1, We report the average discovery loss of the \fixedObj{} object in~\cref{fig:exp2}. We observe that all variants of NDIGO perform better than the baselines by a clear margin. Though ICM performance is not far behind NDIGO (less than two times worse than NDIGO-4). We also observe no significant difference between the different variants of NDIGO in this case. We also report in \cref{tab:exp2} the first visit and visit counts for the \fixedObj{} object and the \tvObj{} object in an episode. NDIGO again demonstrates a superior performance to the baselines. We also observe that NDIGO in most case is not attracted towards the \tvObj{} object. An interesting observation is that, as we increase the horizon of prediction in NDIGO, it takes more time for the agent to find the \fixedObj{} object but at the same time the visit counts increases as well, i.e, the agent stay close to the object for longer time after the first visit.

\begin{figure*}
    \begin{subfigure}[b]{0.32\textwidth}
     \centering
    \includegraphics[width=\textwidth]{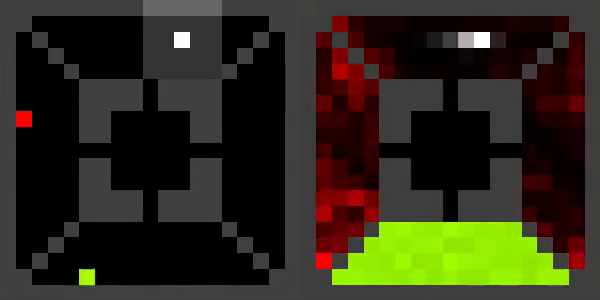}
    \caption{$t=1$}
    \label{fig:exp2.snapshot.a}
    \end{subfigure}
    ~
    \begin{subfigure}[b]{0.32\textwidth}
    \centering
    \includegraphics[width=\textwidth]{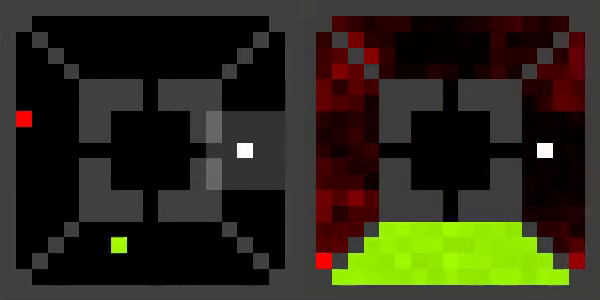}
    \caption{$t=2$}
    \label{fig:exp2.snapshot.b}
    \end{subfigure}
    ~
    \begin{subfigure}[b]{0.32\textwidth}
    \centering
   \includegraphics[width=\textwidth]{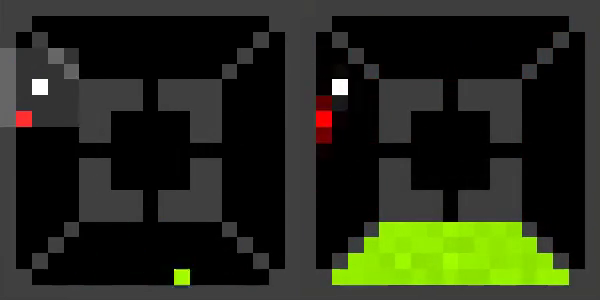}
    \caption{$t=3$}
    \label{fig:exp2.snapshot.c}
   \end{subfigure}
    \caption{Experiment 2: top-down-view snapshots of the behavior of the NDIGO-4 agent. (a) after entering the top-side room (b) after entering the right-side room  (c) after discovering the \fixedObj{} object in the left-side room. In each subpanel the left-side image depicts the ground-truth top-down-view of the world and the right-side image depicts the predicted view from the agent's representation. All times are in seconds.
    \label{fig:exp2.snapshot}}
\end{figure*}

As a qualitative result, we also report top-down-view snapshots of the behavior of NDIGO-$2$ up to the time of discovery of \fixedObj{} in the right room in~\cref{fig:exp2.snapshot}. We also depicts the predicted view of the world from the agent's representation in~\cref{fig:exp2.snapshot}.   As the location of object is unknown to the agent, we observe that the agent searches the top-side, left-side and right-side rooms until it discovers the \fixedObj{} object in the right-side room. It also successfully avoids the bottom-side room containing the \tvObj{} object.  Also as soon as  the agent finds the \fixedObj{} object the uncertainty about the location of \fixedObj{} object completely vanishes (as the agent has learned there is only one \fixedObj{} object exists in the world).  

\begin{table}[ht!]
\scalebox{0.75}{
\begin{tabular}{lcccc}
\toprule
         & \multicolumn{2}{c}{Visit count} & \multicolumn{2}{c}{First visit time} \\
         & \fixedObj{}   & \texttt{w.~noise}  & \fixedObj{}   & \texttt{w.~noise}  \\ 
\midrule
ICM     &     $151.7 \pm 33.0$     &     $15.6 \pm 9.0$     &     $142.1 \pm 40.8$     &     $198.7 \pm 55.1$ \\
NDIGO-1     &     $180.2 \pm 42.7$     &     $12.8 \pm 6.9$     &     $\mathbf{101.1} \pm 31.1$     &     $237.2 \pm 49.4$ \\
NDIGO-2     &     $209.3 \pm 34.9$     &     $\mathbf{3.5} \pm 2.3$     &     $121.1 \pm 36.5$     &     $\mathbf{306.4} \pm 43.4$ \\
NDIGO-4     &     $\mathbf{233.7} \pm 41.6$     &     $5.3 \pm 3.7$     &     $126.7 \pm 43.3$     &     $268.2 \pm 53.1$ \\
\bottomrule
\end{tabular}}

\caption{Average values of the visit counts and first visit time of the trained agent for the \fixedObj{} and \tvObj{} objects in Experiment 2.}

\label{tab:exp2}
\end{table}

\paragraph{Experiment 3.} We investigate whether NDIGO is able to discover and retain the dynamics of moving (but still predictable) objects even when not being in its field of view. For this, we used a \fiverooms{} setting with two \bouncingObj{} objects in upper and lower rooms and a \tvObj{} object in the right room.
\begin{figure}
    \centering
    \includegraphics[width=\columnwidth]{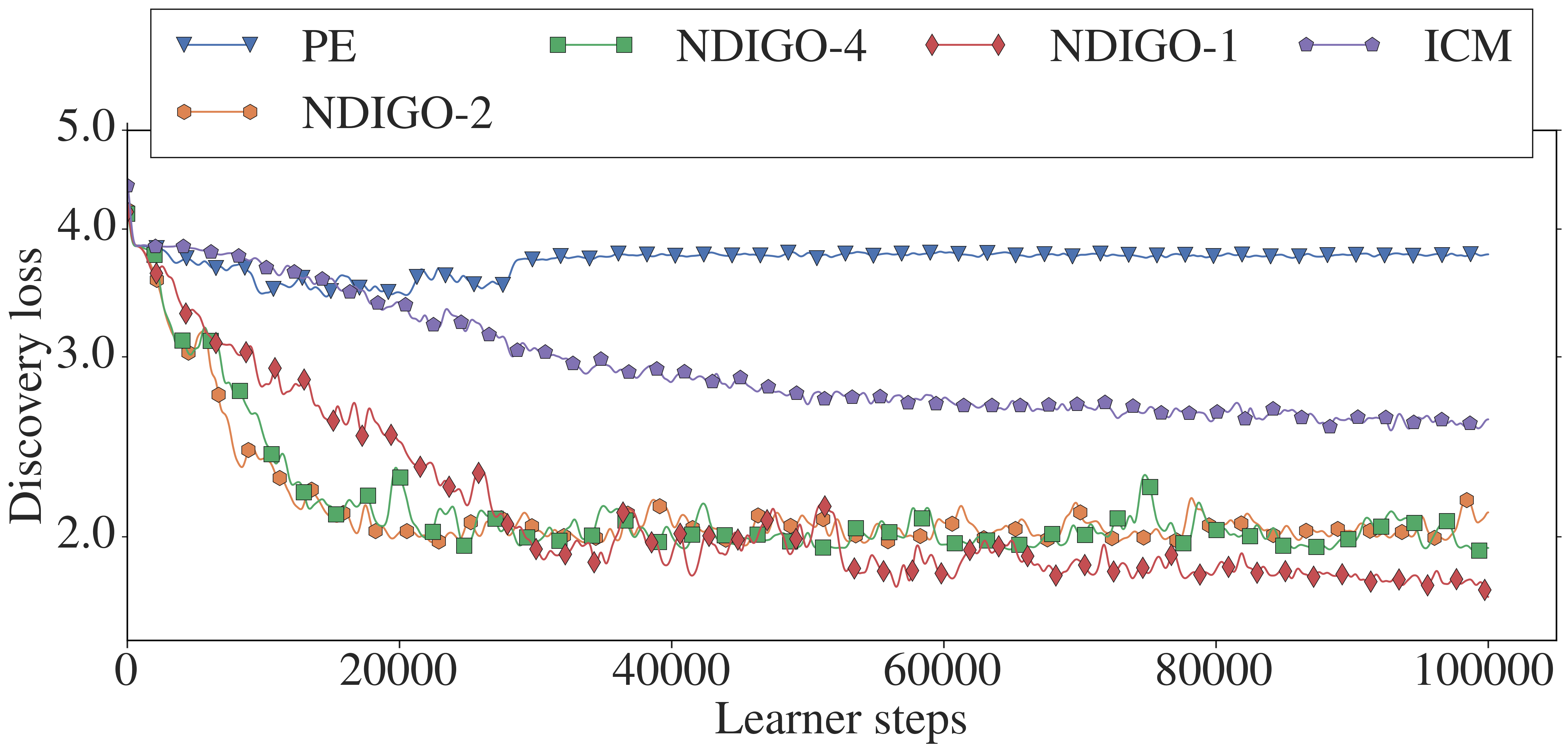}
    \caption{Experiment 3: Average discovery loss of \bouncingObj{} objects. The results are averaged over 10 seeds.
    \label{fig:exp3}}
\end{figure}

\begin{figure}
    \begin{subfigure}[b]{\columnwidth}
    \centering
    \includegraphics[width=0.8\columnwidth]{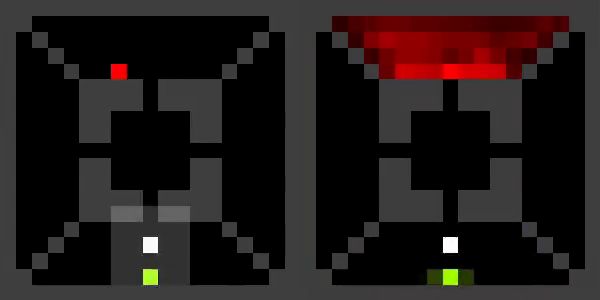}
    \caption{$t=1$}
     \label{fig:exp3.snapshot.a}
    \end{subfigure}
    \hfill
    \begin{subfigure}[b]{\columnwidth}
        \centering
    \includegraphics[width=0.8\columnwidth]{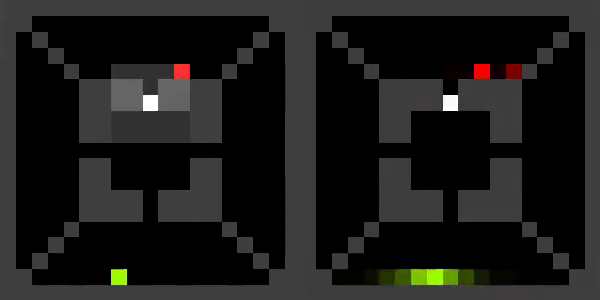}
    \caption{$t=2$}
    \label{fig:exp3.snapshot.b}
    \end{subfigure}
    \caption{Experiment 3: top-down-view snapshots of the behavior of the NDIGO-1 agent. (a) after discovering the green \bouncingObj{} object in the bottom-side room (b)  after discovering the red \bouncingObj{} object in the top-side room. In each subpanel the left-side image depicts the ground-truth top-down-view of the world and the right-side image depicts the predicted view from the agent's representation. All times are in seconds.
    \label{fig:exp3.snapshot}}
\end{figure}

\begin{table}[ht!]
\scalebox{0.75}{
\begin{tabular}{lcccc}
\toprule
         & \multicolumn{2}{c}{Visit count} & \multicolumn{2}{c}{First visit time} \\
         & upper obj. & lower obj. &  upper obj.   & lower obj.  \\ 
\midrule
ICM     &     $80.5 \pm 28.3$     &     $\mathbf{89.1} \pm 28.6$     &     $174.8 \pm 53.4$     &     $127.8 \pm 51.4$ \\
NDIGO-1     &     $41.0 \pm 8.5$     &     $45.2 \pm 11.6$     &     $\mathbf{34.4} \pm 18.7$     &     $\mathbf{38.8} \pm 16.1$ \\
NDIGO-2     &     $108.5 \pm 25.1$     &     $31.3 \pm 20.9$     &     $118.3 \pm 50.4$     &     $312.6 \pm 50.6$ \\
NDIGO-4     &     $\mathbf{198.7} \pm 33.4$     &     $44.2 \pm 28.8$     &     $64.5 \pm 38.8$     &     $320.8 \pm 47.5$ \\
\bottomrule
\end{tabular}}
\caption{Average values of the visit counts and first visit time of the trained agent for the \bouncingObj{} objects in Experiment 3.}

\label{tab:exp3}
\end{table}
We report the discovery loss in~\cref{fig:exp3}. We observe that all variants of NDIGO outperforms the baselines by a large margin in terms of the discovery loss of the \bouncingObj{} object. As the discovery loss for both \bouncingObj{} objects is small, this indicates that NDIGO can encode the dynamics of \bouncingObj{} objects in its representation. We report the first-visit and visit counts for the \bouncingObj{} objects in~\cref{tab:exp3}. NDIGO has a superior performance than the baselines both in terms of visit counts and visit time to the \fixedObj{} objects except for the visit count of the lower object in which ICM produces the best performance. Finally, as a qualitative result, we also report top-down-view snapshots of the behavior of NDIGO-$1$ after the discovery of each \bouncingObj{} object in~\cref{fig:exp3.snapshot}. We observe that the agent can estimate the location of both \bouncingObj{}s in the first visit. Also after departing from the green \bouncingObj{} object and moving towards the red \bouncingObj{} object, still it can track the dynamics of the green \bouncingObj{} object with some small error. This is despite the fact that the green \bouncingObj{} object is not anymore observed by the agent. 

\paragraph{Experiment 4.} We investigate if the horizon $H$ affects the performance of the agents in terms of its sensitivity to structured noise. For this we evaluated which objects the agent seeks in a \fiverooms{} setting with a \brownianObj{} object in the upper room and a \fixedObj{} object in the lower room. In the upper room, the \brownianObj{} moves at every time step. For the \brownianObj{}, unlike \tvObj{}, it is  not guaranteed that the reward of NDIGO is zero. However by increasing the horizon, one may expect that the intrinsic reward due to the \brownianObj{} object becomes negligible because it becomes harder to predict with higher horizons.  

\begin{figure}
    \centering
    \includegraphics[width=\columnwidth]{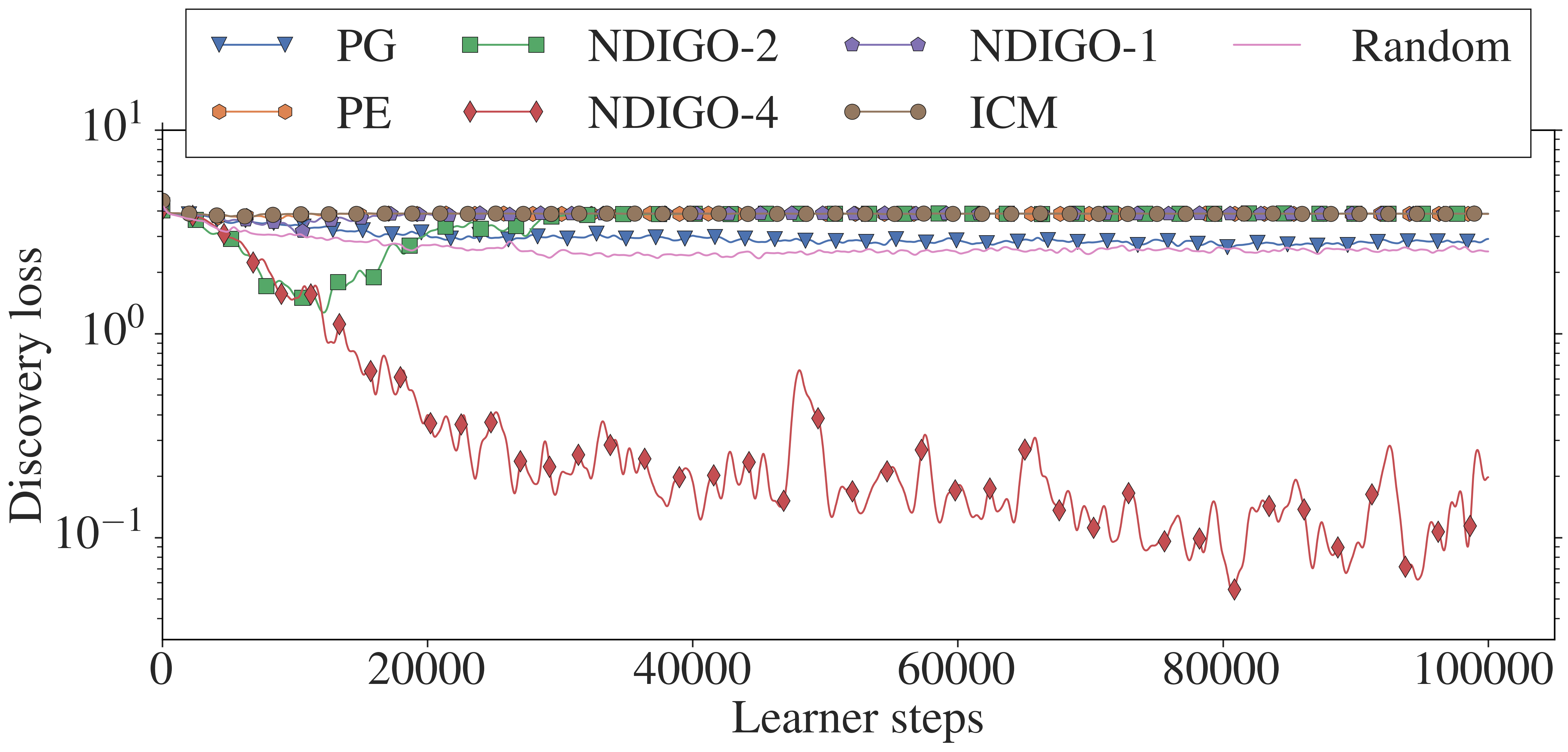}
    \caption{Experiment 4: Average discovery loss of the \fixedObj{} object . The results are averaged over 10 seeds.
    \label{fig:exp4}}
\end{figure}

\begin{table}[ht!]
\scalebox{0.7}{
\centering
\begin{tabular}{lcccc}
\toprule
         & \multicolumn{2}{c}{Visit count} & \multicolumn{2}{c}{First visit time} \\
         & \brownianObj{} & \fixedObj{} &  \brownianObj{}   &  \fixedObj{}  \\ 
\midrule
ICM     &     $358.3 \pm 9.4$     &     $0.5 \pm 0.9$     &     $34.0 \pm 8.3$     &     $385.1 \pm 24.6$ \\
NDIGO-1     &     $356.1 \pm 6.9$     &     $0.0 \pm 0.0$     &     $23.4 \pm 6.4$     &     $398.9 \pm 8.9$ \\
NDIGO-2     &     $350.7 \pm 5.4$     &     $0.1 \pm 0.3$     &     $21.1 \pm 4.8$     &     $383.9 \pm 25.6$ \\
NDIGO-4     &     $\mathbf{0.4} \pm 1.0$     &     $\mathbf{290.5} \pm 31.4$     &     $\mathbf{395.5} \pm 12.4$     &     $\mathbf{68.4} \pm 29.8$ \\
\bottomrule
\end{tabular}
}
\caption{Average values of the visit counts and first visit time of the trained agent for the \brownianObj{} and \fixedObj{} objects in Experiment 4, with all baselines.}
\label{tab:exp4}

\end{table}

\begin{figure}[ht!]
    \centering
    \includegraphics[width=\columnwidth]{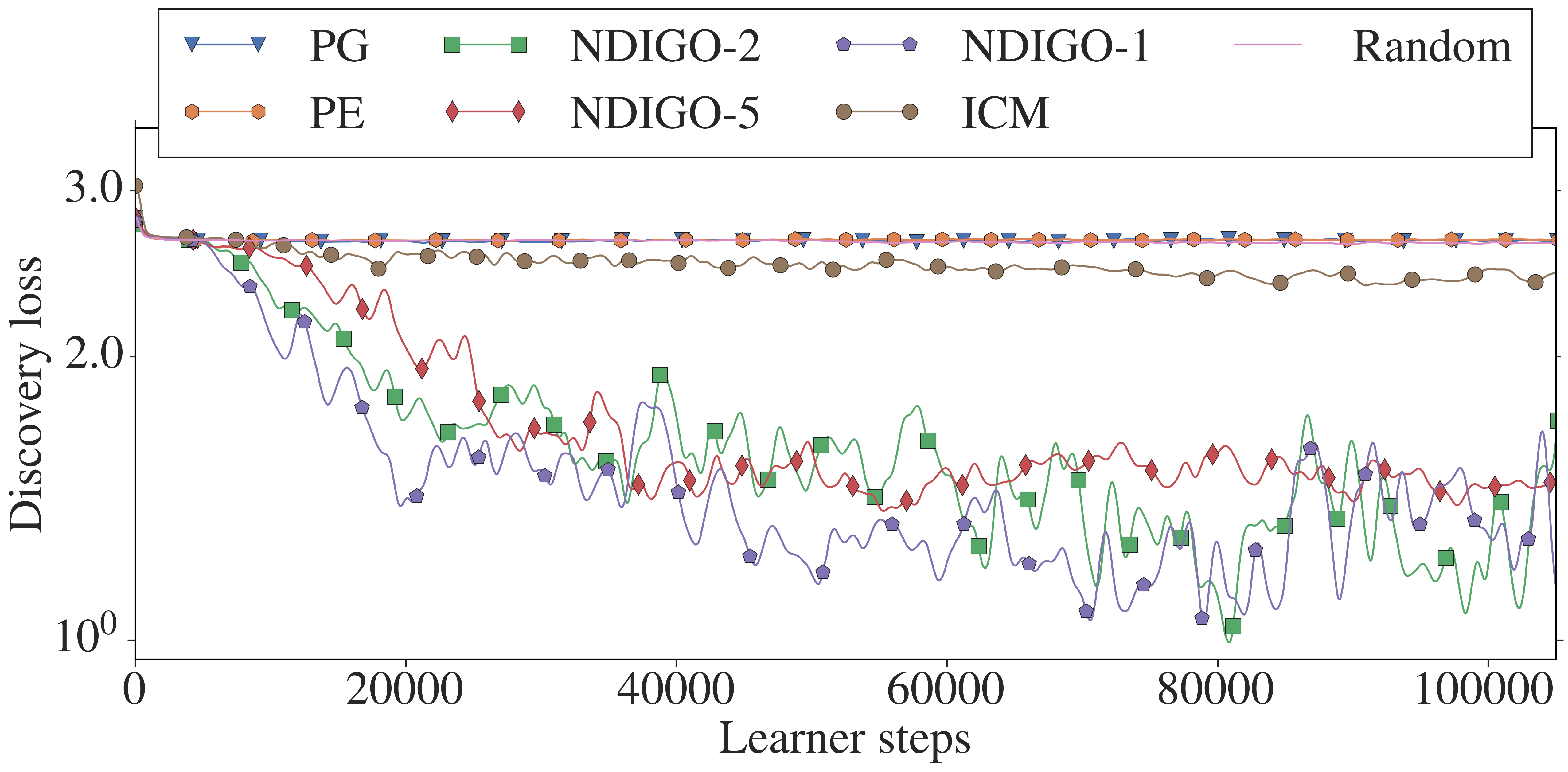}
    \caption{Experiment 5: Average  discovery loss of the \fixedObj{}  and \movableObj{} objects. The results are averaged over 10 seeds.
    \label{fig:exp5}}
\end{figure}

\begin{figure*}
    \centering
    \begin{subfigure}[b]{0.49\textwidth}
    \includegraphics[width=\textwidth]{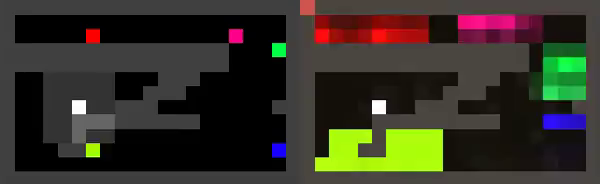}
    \caption{$t=0$}
    \label{fig:exp5.snapshot.a}
    \end{subfigure}
    ~
    \begin{subfigure}[b]{0.49\textwidth}
    \includegraphics[width=\textwidth]{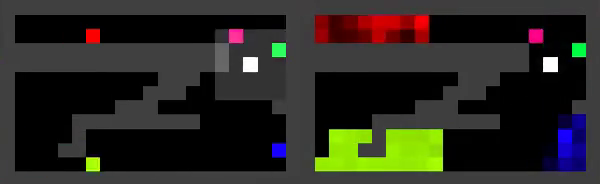}
    \caption{$t=1$}
    \label{fig:exp5.snapshot.b}
    \end{subfigure}
    \hfill
    \begin{subfigure}[b]{0.49\textwidth}
    \includegraphics[width=\textwidth]{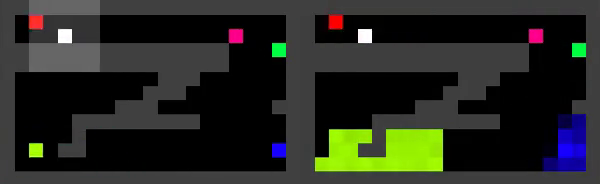}
    \caption{$t=2$}
    \label{fig:exp5.snapshot.c}
    \end{subfigure}
    ~
    \begin{subfigure}[b]{0.49\textwidth}
    \includegraphics[width=\textwidth]{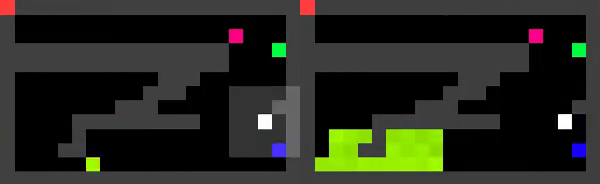}
    \caption{$t=4$}
    \label{fig:exp5.snapshot.d}
    \end{subfigure}
    \caption{Experiment 5: top-down-view snapshots of the behavior of the NDIGO-1 agent in the maze  problem: (a) at the beginning of the episode (b) after discovering the \fixedObj{} objects in room 3 and 4  (c) after discovering the \movableObj{} object in room 5 (d) after discovering the \fixedObj{} object in room 2.  In each subpanel the left-side image depicts the ground-truth top-down-view of the world and the right-side image depicts the predicted view from the agent's representation. All times are in seconds.
    \label{fig:exp5.snapshot}}
\end{figure*}

\begin{table*}[ht!]
\centering
\begin{tabular}{lccccc}
\toprule
         & \multicolumn{5}{c}{Visit frequency} \\
         & Room 1 & Room 2 & Room 3 & Room 4 & Room 5  \\ 
         & \tvObj{} & \fixedObj{} & \fixedObj{} & \fixedObj{} & \movableObj{} \\
\midrule
ICM     &     $100.0\% \pm 0.0\%$     &     $26.8\% \pm 25.7\%$     &     $13.8\% \pm 20.0\%$     &     $6.5\% \pm 14.3\%$     &     $-$ \\
NDIGO-1     &     $94.7\% \pm 12.9\%$     &     $66.4\% \pm 27.4\%$     &     $71.7\% \pm 26.1\%$     &     $70.4\% \pm 26.4\%$     &     $67.8\% \pm 27.1\%$ \\
NDIGO-2     &     $\mathbf{100.0\%} \pm 0.0\%$     &     $78.3\% \pm 23.9\%$     &     $84.8\% \pm 20.9\%$     &     $\mathbf{83.7\%} \pm 21.4\%$     &     $\mathbf{81.5\%} \pm 22.5\%$ \\
NDIGO-5     &     $100.0\% \pm 0.0\%$     &     $49.6\% \pm 29.0\%$     &     $47.4\% \pm 28.9\%$     &     $18.8\% \pm 22.6\%$     &     $-$ \\
NDIGO-10     &     $100.0\% \pm 0.0\%$     &     $\mathbf{84.1\%} \pm 21.4\%$     &     $\mathbf{95.5\%} \pm 12.2\%$     &     $45.5\% \pm 29.1\%$     &     $-$ \\
\bottomrule
\end{tabular}
\caption{Average frequency of visits to each room for the trained agents.\label{tab:visit-freq-exp5}}
\end{table*}

\begin{table*}[ht!]
\centering
\begin{tabular}{lccccc}
\toprule
         & \multicolumn{5}{c}{First visit time} \\
         & Room 1 & Room 2 & Room 3 & Room 4 & Room 5  \\ 
         & \tvObj{} & \fixedObj{} & \fixedObj{} & \fixedObj{} & \movableObj{} \\
\midrule
ICM     &     $\mathbf{4.4} \pm 3.0$     &     $324.7 \pm 79.5$     &     $375.0 \pm 44.0$     &     $391.8 \pm 24.3$     &     - \\
NDIGO-1     &     $40.6 \pm 57.2$     &     $203.0 \pm 90.6$     &     $190.5 \pm 86.0$     &     $199.9 \pm 85.2$     &     $212.7 \pm 83.2$ \\
NDIGO-2     &     $12.9 \pm 10.7$     &     $171.5 \pm 79.5$     &     $159.4 \pm 68.8$     &     $\mathbf{174.5} \pm 68.9$     &     $\mathbf{192.8} \pm 68.9$ \\
NDIGO-5     &     $6.8 \pm 11.5$     &     $245.1 \pm 94.1$     &     $255.9 \pm 91.4$     &     $344.9 \pm 68.7$     &     - \\
NDIGO-10     &     $8.6 \pm 5.9$     &     $\mathbf{128.0} \pm 75.8$     &     $\mathbf{119.1} \pm 53.4$     &     $283.1 \pm 81.4$     &     - \\
\bottomrule
\end{tabular}
\caption{Average time of first visit to each room for the trained agents.\label{tab:first-visit-exp5}}
\end{table*}

We report the results in~\cref{fig:exp4}. We observe that the ICM baseline as well as the variants of NDIGO with the short horizon are being attracted to the structured randomness generated by the \brownianObj{} object. Only NDIGO-$4$ can ignore the  \brownianObj{} object and discover the fixed object. As a result NDIGO-$4$ is the only algorithm capable of minimising the discovery loss of the \fixedObj{} object.      

\paragraph{Experiment 5.} We now compare discovery ability of the agents in a complex \maze{} environment (see \cref{fig:maze}) with no extrinsic reward. Here, the agent starts in a fixed position in the \maze{} environment, and is given no incentive to explore but its intrinsic reward. This setting is challenging for discovery and exploration, since to go the end of the maze the agents need to take a very long and specific sequence of actions. This highlights the importance of intrinsic rewards that encourage discovery. We report the learning curves of NDIGO as well as the baselines in \cref{fig:exp5}. We observe that in this case different variants of NDIGO outperform the baselines by a wide margin in terms of discovery loss, while NDIGO-1 and NDIGO-2 outperforming NDIGO-5. Note that due to the presence of \movableObj{} object, which is unpredictable upon re-spawning, the average loss in this experiment is higher than the prior \fixedObj{} object experiments. We also evaluate the discovery performance of the agent as the number of rooms it is capable of exploring within the duration of the episode.  We present the average visit frequency and first visit time of each room for the trained agents (see \Cref{tab:first-visit-exp5,tab:visit-freq-exp5}). NDIGO-$1$ and NDIGO-$2$ appear as the only agents capable of reaching the final room, whereas NDIGO-$4$ explores $4$ out of $5$. The rest can not go beyond the \tvObj{}~object.

As a qualitative result, we also report top-down-view snapshots of the behavior of NDIGO-$1$ up to the time of discovery of the last \fixedObj{} object in room 2 in~\cref{fig:exp5.snapshot}. We also depicts the predicted view of the world from the agent's representation in~\cref{fig:exp2.snapshot}. 
We observe the agent drives across the \maze{} all the way from room 1 to room 5 and in the process discovers the \fixedObj{} objects in rooms 3-4 (see \cref{fig:exp2.snapshot.a}) and the movable object in room 5 (see \cref{fig:exp2.snapshot.c}). It then chases \movableObj{} object until \movableObj{} object gets fixated on the top-left corner of the world. The agent then moves back to room 2 (see \cref{fig:exp2.snapshot.c}) and discovers the last blue \fixedObj{} object there, while maintaining its knowledge of the other objects.The reason for ignoring the blue \fixedObj{} object in room 2, in the first place, might be due to the fact that the agent can obtain more intrinsic rewards by chasing the \movableObj{} object. So it tries to reach to room 5 as fast as possible at the expense of ignoring the blue \fixedObj{} object in room 2.

\section{Conclusion}

We aimed at building a proof of concept for a world discovery model by developing the NDIGO agent and comparing its performance with the state-of-the-art information-seeking algorithms in terms of its ability to discover the world.  Specifically, we considered a variety of simple local-view 2D navigation tasks with some hidden randomly-placed objects and looked at whether the agent can discover its environment and the location of objects. We evaluate the ability of our agent for discovery through the glass-box approach which measures how accurate location of objects can be predicted from the internal representation. Our results showed that in all these tasks  NDIGO produces an effective  information seeking strategy capable of discovering the hidden objects without being distracted by the white noise, whereas the baseline information seeking methods in most cases failed to discover the objects due to the presence of noise.                      

There remains much interesting future work to pursue. The ability of our agent to discover its world can be very useful in improving performance in multi-task and transfer settings as the  NDIGO model can be used to discover the the new features of new tasks. Also in this paper we focused on visually simple tasks. To scale up our model to more complex visual tasks we need to consider more powerful prediction models such as Pixel-CNN~\citep{van2016conditional}, VAE~\citep{kingma2013auto}, Info-GAN~\citep{chen2016infogan} and Draw~\citep{gregor2015draw} capable of providing high accuracy predictions for high-dimensional visual scenes. We also can go beyond predicting only visual observations to other modalities of sensory inputs, such as proprioception and touch sensors~\citep{amos2018learning}.

\label{sec:conclusion}

\section*{Acknowledgements}
We would like to thank Daniel Guo, Theophane Webber, Caglar Gulcehre, Toby Pohlen, Steven Kapturovski and Tom Stepleton for insightful discussions, comments and feedback on this work.
\bibliography{biblio}
\bibliographystyle{icml2019}
\newpage
\appendix
\onecolumn

\section{NDIGO Global Network Architecture}

\begin{centering}
\begin{figure}[ht!]
\includegraphics[width=.9\textwidth]{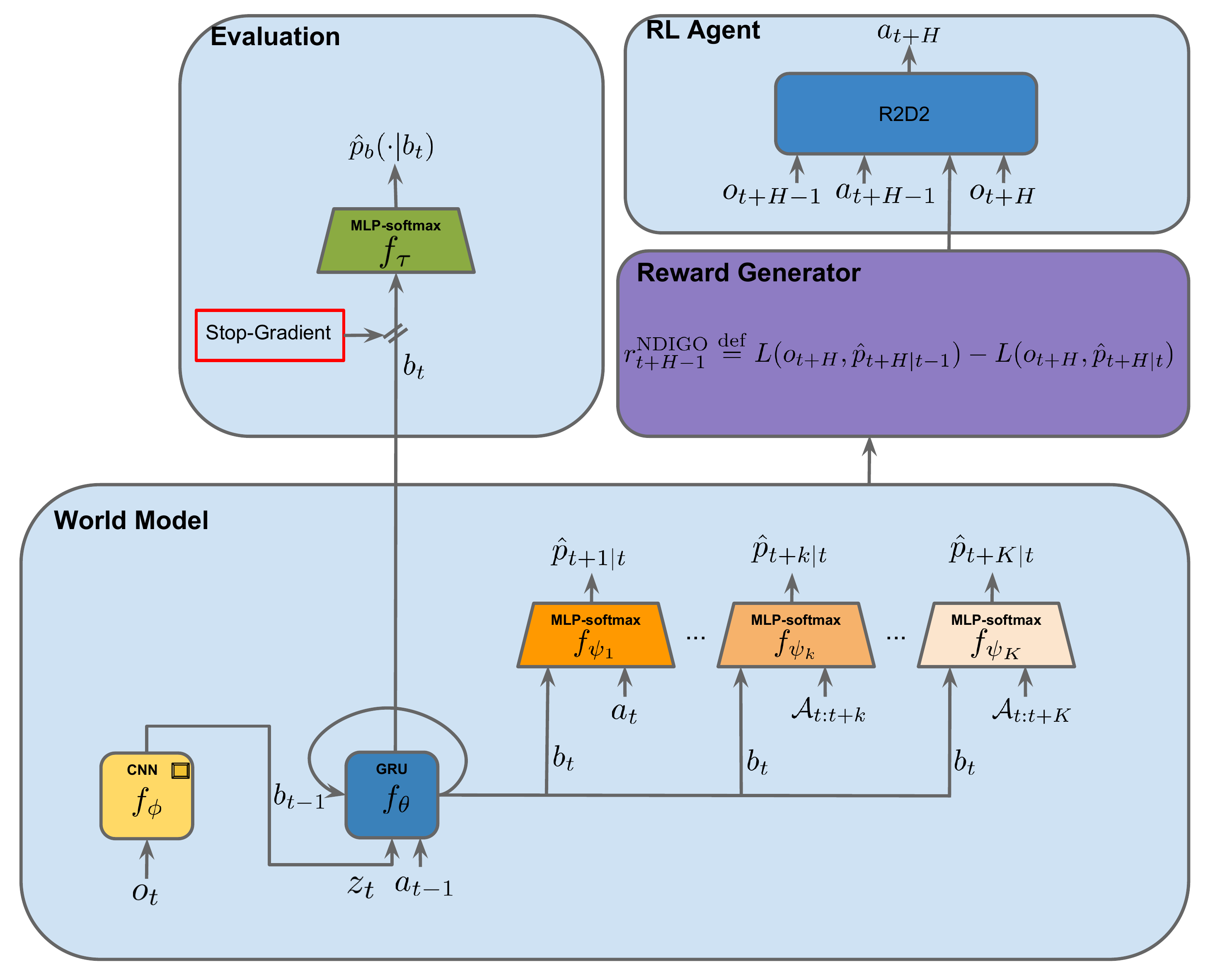}
    \caption{Global Architecture of the NDIGO agent}
    \label{fig:globalNDIGOarchitecture}
\end{figure}
\end{centering}

\section{NDIGO Agent Implementation Details}
\label{sec:agentDetails}

\subsection{World Model}
\label{subsec:worldmodelappendix}
\begin{itemize}
    \item Convolution Neural Network (CNN) $f_{\phi}$: Observations $o_t$ are fed through a two-layer CNN ($16$ $3 \times 3$ filters with $1 \times 1$ stride, then $16$ $3 \times 3$ filters with $2 \times 2$ stride; edges are padded if needed), then through a fully connected single layer perceptron with 256 units, then through a ReLU activation, resulting in a transformed observation $z_t$.
    \item Gated Recurrent Unit (GRU) $f_{\theta}$ : $128$ units GRU.
    \item Frame predictors $\{f_{\psi_k}\}_{k=1}^K$ are MultiLayer Perceptrons (MLPs): one hidden-layer of $64$ units followed by a ReLU activation and the output layer of $25$ units ($5\times 5$ is the size of the local view which the size of the observation $o_t$) followed by a ReLU activation.
    \item $K = 10$
    \item Optimiser for frame predictions: Adam optimiser~\citep{kingma2014adam} with batch size $256$ and learning rate $\num{5e-4}$.  
\end{itemize}

\subsection{Reward Generator}
\begin{itemize}
    \item The horizon $H =\{1, 2, 4\}$ can take one of these values in our experiments.
\end{itemize}

\subsection{Evaluation}

\begin{itemize}
    \item The MLP $f_\tau$: one hidden-layer of $64$ units followed by a ReLU activation and the output layer of $361$ units ($19\times 19$ is the size of the global view of the \fiverooms{} environment which is also the size of the real state $x_t$) followed by a ReLU activation.
    \item Optimiser for evaluation: Adam optimiser~\citep{kingma2014adam} with batch size $256$ and learning rate $\num{5e-4}$. 
\end{itemize}

\subsection{RL Agent}
We use the Recurrent Replay Distributed DQN (R2D2)~\citep{kapturowski2018recurrent} with the following parameters:
\begin{itemize}
    \item Replay: replay period is $100$, replay trace length is $100$, replay size is $\num{1e6}$ and we use uniform prioritisation.
    \item Network architecture: R2D2 uses a two-layer CNN, followed by a GRU which feeds into the advantage and value heads of a dueling network~\citep{wang2015dueling}, each with a hidden layer of size $128$ units. The CNN and GRU of the RL agent have the same architecture and parameters as the one described for the World Model (see Sec.\ref{subsec:worldmodelappendix}) but do not share the same weights.
    \item Algorithm: Retrace Learning update~\citep{munos2016safe} with discount factor $\gamma=0.99$ and eligibility traces coefficient $\lambda=0.97$, target network with update period $1024$, no reward clipping and signed-hyperbolic re-scaling~\citep{pohlen2018observe}.
    \item Distributed training: $100$ actors and $1$ learner, actor update period every $100$ learner steps.
    \item Optimiser for RL: Adam optimiser~\citep{kingma2014adam} with batch size $256$ and learning rate $\num{1e-4}$.
    \item The intrinsic rewards are provided directly to the RL agent without any scaling. 
\end{itemize}

\subsection{Training loop pseudocode}
\label{appendix:ndigo-alg}
\begin{algorithm}
    \caption{NDIGO training loop.}
    \label{alg:ndigo}
    \begin{algorithmic}[1] % The number tells where the line numbering should start
        \REQUIRE{Policy $\mathbf \pi$, history $h_T$, $K \geq H \geq 1$, weights $\mathbf W$}
        \STATE $b_0 \gets 0$
         \FOR{$t=1\dots T-K$} 
            \STATE $z_t = f_\phi(o_t; \mathbf W)$  \COMMENT{Observation CNN}
            \STATE $b_t \gets f_\theta(z_t, a_{t-1}, b_{t-1}; \mathbf W)$  \COMMENT{Belief GRU}
            \STATE $\mathcal A \gets [a_{t-1}]$
            \FOR{$k=1\dots K$}
                \STATE $\hat p_{t+k|t} \gets f_{\psi_k}(b_t, \mathcal A; \mathbf W)$ \COMMENT{Prediction MLP}
                \STATE $L_{t+k|t} \gets -\ln{\hat p_{t+k|t}(o_{t+k})}$
                \STATE $\mathcal A \gets \mathcal A + [a_{t+k-1}]$
            \ENDFOR
            \STATE $r^\mathrm{NDIGO}_{t+H-1} \gets L_{t+H|t-1} - L_{t+H|t}$
        \ENDFOR
        \STATE Update $\mathbf W$ to minimise $\sum_{t=1}^{T-K} \sum_{k=1}^{K} L_{t+k|t}$
        \STATE Update $\mathbf \pi$ using the set of rewards  $\left\{ r_{t+H-1}^\mathrm{NDIGO \vphantom |} \right\}_{t \in 1:T-K+1}$ with the RL Algo.
    \end{algorithmic}
\end{algorithm}

\section{NDIGO Alternative Architecture}

An alternative architecture for NDIGO consists in encoding the sequence of actions $\A_{t:t+k}$ into a representation using a GRU $f_\xi$.
The hidden state of this GRU is $a_{t, k} = f_\psi(a_{t+k}, a_{t, k-1})$, with the initialisation $a_{t, 1} = f_\xi(a_t, 0)$.
Then we use a single neural network $f_{\psi}$ to output, for any $k$, the probability distribution $\hat{p}_{t,k}(.|b_t, a_{t, k})$ when given the input $a_{t, k}$ concatenated with $b_t$. The loss function for the network $f_{\psi}$ at time step $t+k-1$ is a cross entropy loss:

\begin{align}
L(o_{t+k}, \hat{p}_{t, k}(.|b_t, a_{t, k}))= - \ln (\hat{p}_{t, k}(o_{t+k}|b_t, a_{t, k})).
\end{align}

\begin{figure}[ht!]
\begin{centering}
\includegraphics[width=.7\textwidth]{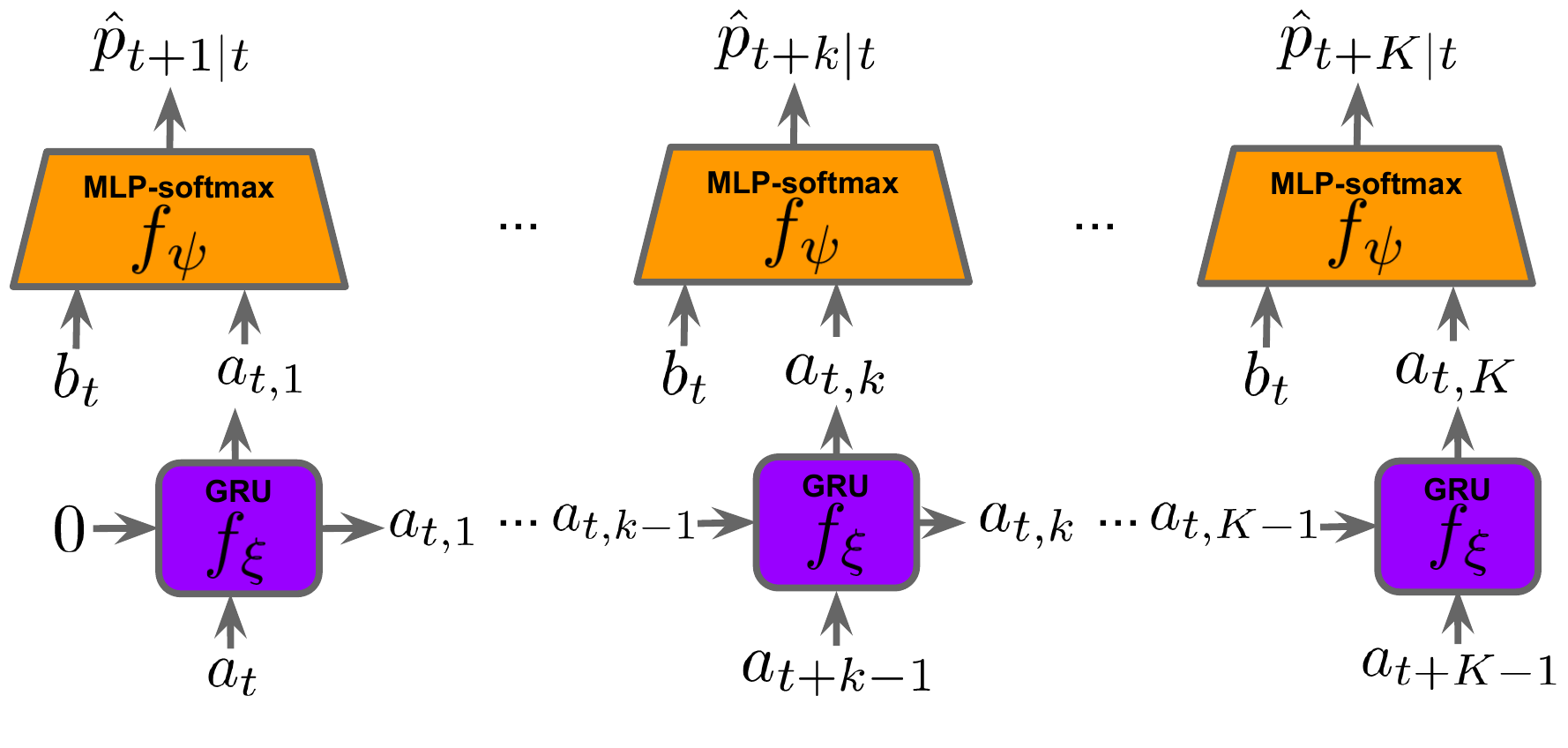}
    \caption{Alternative architecture of NDIGO for the frame prediction tasks.}
    \label{fig:framepredictor}
\end{centering}
\end{figure}

\section{\citet{pathak2017curiosity}'s ICM Model for Partially Observable Environments}
\label{appendix:icm}
\begin{figure}[ht!]
\begin{centering}
\includegraphics[width=.7\textwidth]{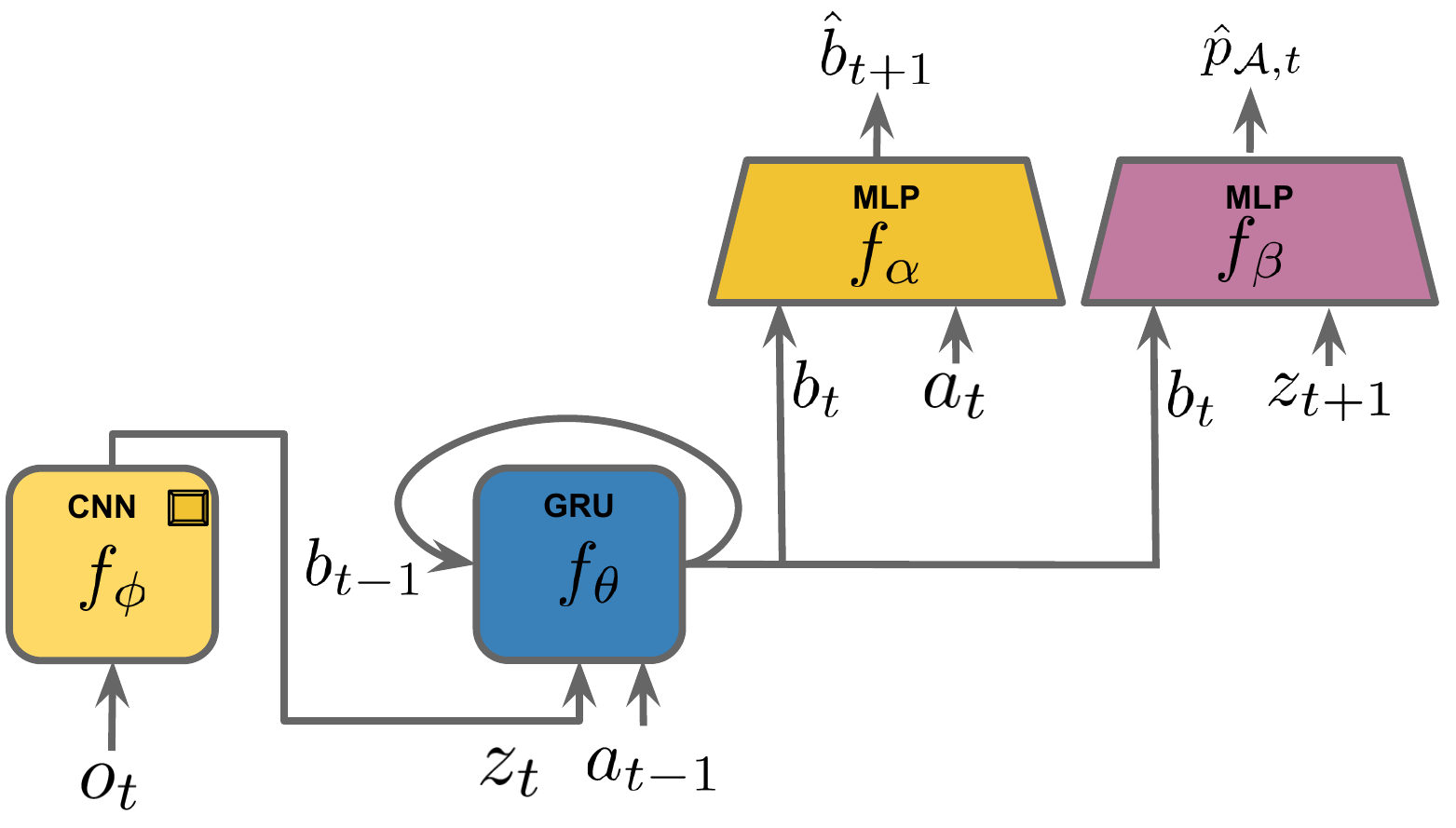}
    \caption{\citet{pathak2017curiosity}'s ICM Model for Partially Observable Environments}
    \label{fig:ICMmodel}
\end{centering}
\end{figure}

The method consists in training the internal representation $b_t$ to be less sensitive to noise using a self-supervised inverse dynamics model.
To do so, one inverse dynamics model $f_{\beta}$ fed by $(b_{t}, z_{t+1})$ (concatenation of the internal representation and the transformed observation $z_{t+1}$)  outputs a distribution $\hat{p}_{\A}$ over actions that predicts the action $a_{t}$. This network is trained by the loss: $L(\hat{p}_{\A}, a_t)= -\ln(\hat{p}_{\A}(a_t))$. Then a forward model $f_{\beta}$ fed by $(b_t, a_t)$ (concatenation of the internal representation and the action) outputs a vector $\hat{b}_{t+1}$ that directly predict the future internal representation $b_{t+1}$. The forward model $f_{\alpha}$ is trained with a regression loss: $L_2(\hat{b}_{t+1},b_{t+1})= \|\hat{b}_{t+1}-b_{t+1}\|^2_2$. The neural architecture is shown in Fig.~\ref{fig:ICMmodel}. Finally, the intrinsic reward is defined as:    
\begin{equation}
   r^{\mathrm{FPE}}_{t} = L_2(\hat{b}_{t+1},b_{t+1}).
\end{equation}
This is slightly different from the architecture proposed by~\citet{pathak2017curiosity} in order to be compatible with partially observable environments.

\subsection{Details of the ICM Model's Architecture}
The ICM agent shares exactly the same architecture than the NDIGO agent except that the forward predictors $\{f_{\psi_k}\}_{k=1}^K$ are replaced by an inverse model $f_{\beta}$ and a forward model $f_{\alpha}$.

\begin{itemize}
    \item The inverse model $f_{\beta}$ is an MLP: one hidden layer of $256$ units and the output layer of $5$ units (one-hot action size).
    \item The forward model $f_{\alpha}$ is an MLP: one hidden layer of $256$ units and the output layer of $128$ units (size of the GRU).
\end{itemize}

\section{Additional results}

\subsection{Additional results for Experiment 2-4}
\label{subsec:additionalexp2-4}
\Cref{tab:exp2-extra,tab:exp3-extra,tab:exp4-extra} contain the results (including baselines) for experiments 2 to 4.

\begin{table}[ht!]
\centering
\begin{tabular}{lcccc}
\toprule
         & \multicolumn{2}{c}{Visit count} & \multicolumn{2}{c}{First visit time} \\
         & \fixedObj{}   & \tvObj{}  & \fixedObj{}   & \tvObj{}  \\ 
\midrule
Random     &     $20.8 \pm 16.8$     &     $51.0 \pm 15.8$     &     $332.3 \pm 42.4$     &     $148.5 \pm 56.9$ \\
PE     &     $0.3 \pm 0.8$     &     $161.7 \pm 3.4$     &     $388.8 \pm 21.0$     &     $11.2 \pm 2.5$ \\
PG     &     $32.6 \pm 26.6$     &     $11.7 \pm 7.7$     &     $309.9 \pm 49.9$     &     $293.2 \pm 50.9$ \\
ICM     &     $151.7 \pm 33.0$     &     $15.6 \pm 9.0$     &     $142.1 \pm 40.8$     &     $198.7 \pm 55.1$ \\
NDIGO-1     &     $180.2 \pm 42.7$     &     $12.8 \pm 6.9$     &     $\mathbf{101.1} \pm 31.1$     &     $237.2 \pm 49.4$ \\
NDIGO-2     &     $209.3 \pm 34.9$     &     $\mathbf{3.5} \pm 2.3$     &     $121.1 \pm 36.5$     &     $\mathbf{306.4} \pm 43.4$ \\
NDIGO-4     &     $\mathbf{233.7} \pm 41.6$     &     $5.3 \pm 3.7$     &     $126.7 \pm 43.3$     &     $268.2 \pm 53.1$ \\
\bottomrule
\end{tabular}
\caption{Average values of the visit counts and first visit time of the trained agent for the \fixedObj{} and \tvObj{} objects in Experiment 2, with all baselines.}
\label{tab:exp2-extra}
\end{table}

\begin{table}[ht!]
\centering
\begin{tabular}{lcccc}
\toprule
         & \multicolumn{2}{c}{Visit count} & \multicolumn{2}{c}{First visit time} \\
         & upper obj. & lower obj. &  upper obj.   & lower obj.  \\ 
\midrule
PE     &     $0.6 \pm 0.5$     &     $0.1 \pm 0.2$     &     $343.8 \pm 42.6$     &     $390.3 \pm 18.2$ \\
ICM     &     $80.5 \pm 28.3$     &     $\mathbf{89.1} \pm 28.6$     &     $174.8 \pm 53.4$     &     $127.8 \pm 51.4$ \\
NDIGO-1     &     $41.0 \pm 8.5$     &     $45.2 \pm 11.6$     &     $\mathbf{34.4} \pm 18.7$     &     $\mathbf{38.8} \pm 16.1$ \\
NDIGO-2     &     $108.5 \pm 25.1$     &     $31.3 \pm 20.9$     &     $118.3 \pm 50.4$     &     $312.6 \pm 50.6$ \\
NDIGO-4     &     $\mathbf{198.7} \pm 33.4$     &     $44.2 \pm 28.8$     &     $64.5 \pm 38.8$     &     $320.8 \pm 47.5$ \\
\bottomrule
\end{tabular}
\caption{Average values of the visit counts and first visit time of the trained agent for the \bouncingObj{} objects in Experiment 3, with the PE and ICML baselines.}
\label{tab:exp3-extra}
\end{table}

\begin{table}[ht!]
\centering
\begin{tabular}{lcccc}
\toprule
         & \multicolumn{2}{c}{Visit count} & \multicolumn{2}{c}{First visit time} \\
         & \brownianObj{} & \fixedObj{} &  \brownianObj{}   &  \fixedObj{}  \\ 
\midrule
Random     &     $16.7 \pm 11.6$     &     $41.1 \pm 24.7$     &     $309.3 \pm 46.0$     &     $244.3 \pm 56.1$ \\
PE     &     $357.3 \pm 4.6$     &     $0.1 \pm 0.3$     &     $15.6 \pm 3.5$     &     $399.2 \pm 8.5$ \\
PG     &     $23.0 \pm 14.0$     &     $38.2 \pm 25.7$     &     $281.7 \pm 52.4$     &     $268.4 \pm 57.2$ \\
ICM     &     $358.3 \pm 9.4$     &     $0.5 \pm 0.9$     &     $34.0 \pm 8.3$     &     $385.1 \pm 24.6$ \\
NDIGO-1     &     $356.1 \pm 6.9$     &     $0.0 \pm 0.0$     &     $23.4 \pm 6.4$     &     $398.9 \pm 8.9$ \\
NDIGO-2     &     $350.7 \pm 5.4$     &     $0.1 \pm 0.3$     &     $21.1 \pm 4.8$     &     $383.9 \pm 25.6$ \\
NDIGO-4     &     $\mathbf{0.4} \pm 1.0$     &     $\mathbf{290.5} \pm 31.4$     &     $\mathbf{395.5} \pm 12.4$     &     $\mathbf{68.4} \pm 29.8$ \\
\bottomrule
\end{tabular}
\caption{Average values of the visit counts and first visit time of the trained agent for the \brownianObj{} and \fixedObj{} objects in Experiment 4, with all baselines.}
\label{tab:exp4-extra}
\end{table}

\subsection{Additional results for Experiment 5}
\Cref{tab:visit-freq-exp5-full,tab:first-visit-exp5-full} present the complete results of Experiment 5; note that a room is considered as visited when the agent has actually seen the object inside that room. As the object in Room 2 can be missed by the agent if it appears in the lower-right corner of the maze, the reported frequency of visits to Room 2 can be lower than that of Rooms 3 and beyond, as this is the case for the reported figures of the NDIGO-1 and NDIGO-2 agents. A dash symbol indicates that the agent never visits the corresponding room.

\label{appendix:tables-full-exp5}
\begin{table}[ht!]
\centering
\begin{tabular}{lccccc}
\toprule
         & \multicolumn{5}{c}{Visit frequency} \\
         & Room 1 & Room 2 & Room 3 & Room 4 & Room 5  \\ 
         & \tvObj{} & \fixedObj{} & \fixedObj{} & \fixedObj{} & \movableObj{} \\
\midrule
Random     &     $100.0\% \pm 0.0\%$     &     $0.9\% \pm 5.5\%$     &     $-$     &     $-$     &     $-$ \\
PE     &     $100.0\% \pm 0.0\%$     &     $-$     &     $-$     &     $-$     &     $-$ \\
PG     &     $93.6\% \pm 14.3\%$     &     $-$     &     $-$     &     $-$     &     $-$ \\
ICM     &     $100.0\% \pm 0.0\%$     &     $26.8\% \pm 25.7\%$     &     $13.8\% \pm 20.0\%$     &     $6.5\% \pm 14.3\%$     &     $-$ \\
NDIGO-1     &     $94.7\% \pm 12.9\%$     &     $66.4\% \pm 27.4\%$     &     $71.7\% \pm 26.1\%$     &     $70.4\% \pm 26.4\%$     &     $67.8\% \pm 27.1\%$ \\
NDIGO-2     &     $\mathbf{100.0\%} \pm 0.0\%$     &     $78.3\% \pm 23.9\%$     &     $84.8\% \pm 20.9\%$     &     $\mathbf{83.7\%} \pm 21.4\%$     &     $\mathbf{81.5\%} \pm 22.5\%$ \\
NDIGO-5     &     $100.0\% \pm 0.0\%$     &     $49.6\% \pm 29.0\%$     &     $47.4\% \pm 28.9\%$     &     $18.8\% \pm 22.6\%$     &     $-$ \\
NDIGO-10     &     $100.0\% \pm 0.0\%$     &     $\mathbf{84.1\%} \pm 21.4\%$     &     $\mathbf{95.5\%} \pm 12.2\%$     &     $45.5\% \pm 29.1\%$     &     $-$ \\
\bottomrule
\end{tabular}
\caption{Average frequency of visits to each room for the trained agents.\label{tab:visit-freq-exp5-full}}
\end{table}

\begin{table}[ht!]
\centering
\begin{tabular}{lccccc}
\toprule
         & \multicolumn{5}{c}{First visit time} \\
         & Room 1 & Room 2 & Room 3 & Room 4 & Room 5  \\ 
         & \tvObj{} & \fixedObj{} & \fixedObj{} & \fixedObj{} & \movableObj{} \\
\midrule
Random     &     $15.1 \pm 18.6$     &     $399.9 \pm 6.9$     &     -     &     -     &     - \\
PE     &     $8.6 \pm 4.3$     &     -     &     -     &     -     &     - \\
PG     &     $33.9 \pm 56.3$     &     -     &     -     &     -     &     - \\
ICM     &     $\mathbf{4.4} \pm 3.0$     &     $324.7 \pm 79.5$     &     $375.0 \pm 44.0$     &     $391.8 \pm 24.3$     &     - \\
NDIGO-1     &     $40.6 \pm 57.2$     &     $203.0 \pm 90.6$     &     $190.5 \pm 86.0$     &     $199.9 \pm 85.2$     &     $212.7 \pm 83.2$ \\
NDIGO-2     &     $12.9 \pm 10.7$     &     $171.5 \pm 79.5$     &     $159.4 \pm 68.8$     &     $\mathbf{174.5} \pm 68.9$     &     $\mathbf{192.8} \pm 68.9$ \\
NDIGO-5     &     $6.8 \pm 11.5$     &     $245.1 \pm 94.1$     &     $255.9 \pm 91.4$     &     $344.9 \pm 68.7$     &     - \\
NDIGO-10     &     $8.6 \pm 5.9$     &     $\mathbf{128.0} \pm 75.8$     &     $\mathbf{119.1} \pm 53.4$     &     $283.1 \pm 81.4$     &     - \\
\bottomrule
\end{tabular}
\caption{Average time of first visit to each room for the trained agents.\label{tab:first-visit-exp5-full}}
\end{table}

\end{document}